\documentclass[10pt,twocolumn,letterpaper]{article}

\usepackage[final]{iccv}      %

\usepackage{float}
\usepackage{amsmath}

\newcommand{\modelname}{\textsc{Via}}

\newcommand{\gatherswap}{\emph{gather-and-swap}}
\newcommand{\gath}{\emph{gather}}
\newcommand{\swap}{\emph{swap}}

\newcommand{\bs}[1]{{\boldsymbol{#1}}}

\newcommand{\mask}{\mathbf{M}}
\newcommand{\maskt}{\mathbf{M_{t}}}

\newcommand{\softmax}{\mathrm{softmax}}

\definecolor{iccvblue}{rgb}{0.21,0.49,0.74}
\usepackage[pagebackref,breaklinks,colorlinks,allcolors=iccvblue]{hyperref}

\def\paperID{10648} %
\def\confName{ICCV}
\def\confYear{2025}

\title{\modelname{}: Unified Spatiotemporal \underline{Vi}deo \underline{A}daptation for Global and Local Video Editing}

\author{
Jing Gu$^{1}$\quad Yuwei Fang$^{2}$ \quad Ivan Skorokhodov$^{2}$ \quad Peter Wonka$^{3}$ \quad Xinya Du$^{4}$ \\
Sergey Tulyakov$^{2}$ \quad Xin Eric Wang$^{1}$\\ 
\texttt{\textsuperscript{1}University of California, Santa Cruz\; \textsuperscript{2}Snap Research} \\
\texttt{\textsuperscript{3}KAUST\; \textsuperscript{4}University of Texas at Dallas}
\and
\texttt{\{jgu110, xwang366\}@ucsc.edu, yfang3@snapchat.com} \\
\url{https://via-video.github.io/}
}

\begin{document}

\twocolumn[{
\renewcommand\twocolumn[1][]{#1}
\maketitle
\centering
\includegraphics[width=0.95\textwidth]{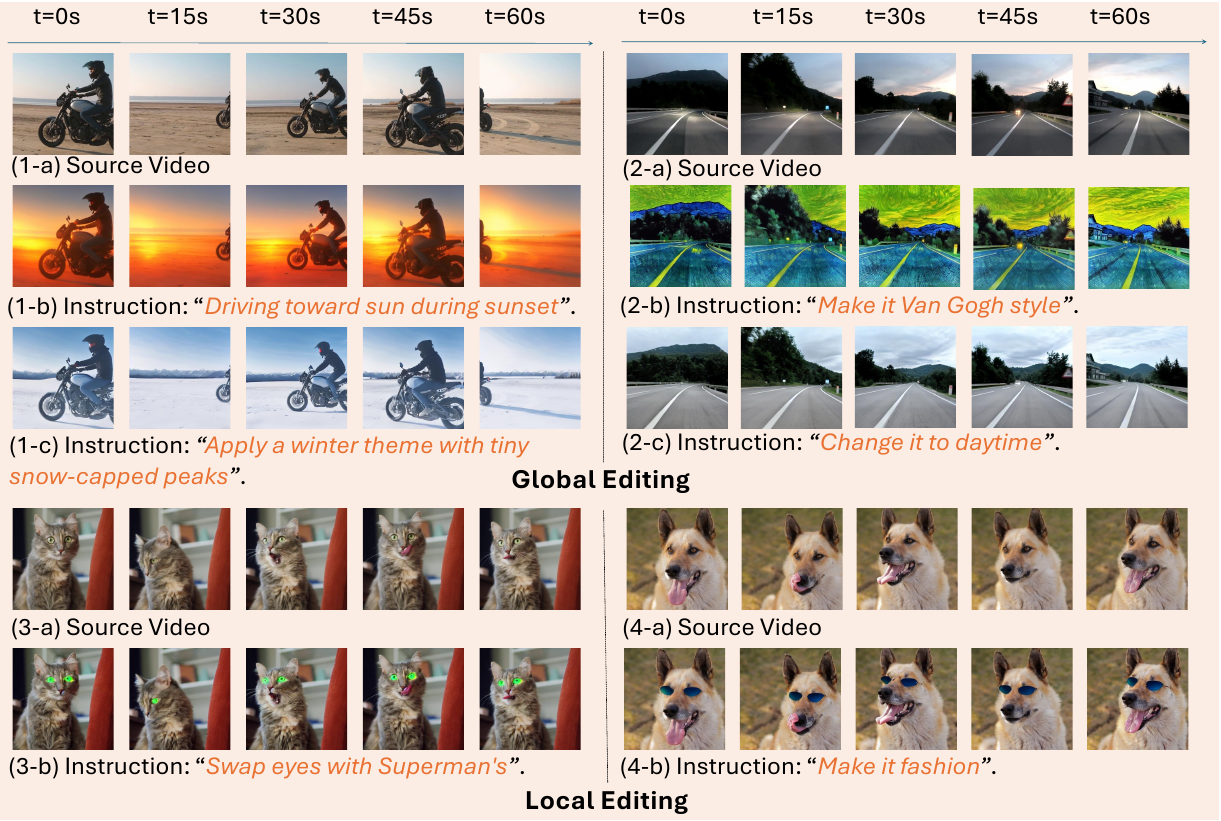}
\hfill\captionof{figure}{ \textbf{Video editing results by \modelname{}. }\modelname{} excels in \textit{precise} and \textit{consistent} editing across diverse video tasks. Top: consistent results over long videos with a duration of 1 minute, which is challenging in current literature. Bottom: consistent results for precise local editing.}

\vspace{1em}
\label{fig:teaser} 
}]

\begin{abstract}

Video editing serves as a fundamental pillar of digital media, spanning applications in entertainment, education, and professional communication.
However, previous methods often overlook the necessity of comprehensively understanding both global and local contexts, leading to inaccurate and inconsistent edits in the spatiotemporal dimension, especially for long videos.
In this paper, we introduce \modelname{}, a unified spatiotemporal \underline{VI}deo \underline{A}daptation framework for global and local video editing, pushing the limits of consistently editing minute-long videos.
First, to ensure local consistency within individual frames, we designed \emph{test-time editing adaptation} to adapt a pre-trained image editing model for improving consistency between potential editing directions and the text instruction, and adapt masked latent variables for precise local control.
Furthermore, to maintain global consistency over the video sequence, we introduce \emph{spatiotemporal adaptation} that recursively \gath{} consistent attention variables in key frames and strategically applies them across the whole sequence to realize the editing effects.
Extensive experiments demonstrate that, compared to baseline methods, our \modelname{} approach produces edits that are more faithful to the source videos, more coherent in the spatiotemporal context, and more precise in local control. More importantly, we show that \modelname{} can achieve consistent long video editing in minutes, unlocking the potential for advanced video editing tasks over long video sequences. 
\end{abstract}

\section{Introduction}
With the exponential growth of digital content creation, video editing has become essential across various domains, including filmmaking~\citep{frierson2018film,dancyger2018technique}, advertising~\citep{mei2007videosense,kholisoh2021short}, education~\citep{calandra2008exploratory,calandra2009using}, and social media~\citep{jackson2016digital,schmitz2006international}. This task presents significant challenges, such as preserving the integrity of the original video, accurately following user instructions, and ensuring consistent editing quality across both time and space. These challenges are particularly pronounced in longer videos, where maintaining long-range spatiotemporal consistency is critical.

A substantial body of research has explored video editing models. One approach uses video models to process the source video as a whole~\citep{ku2024anyv2v, liu2023video}. However, due to limitations in model capacity and hardware, these methods are typically effective only for short videos (fewer than 200 frames). To overcome these limitations, various methods have been proposed~\citep{xing2023simda,wu2023tune,guo2023animatediff,wu2023fairy}. Another line of research leverages the success of image-based models~\citep{ho2022classifier,nichol2022glide,podell2023sdxl,avrahami2022blended,brooks2023instructpix2pix} by adapting their image-editing capabilities to ensure temporal consistency during test time~\citep{khachatryan2023text2video, tokenflow2023,wu2023fairy,qi2023fatezero,wang2023zero}. However, inconsistencies accumulate in this frame-by-frame editing process, causing the edited video to deviate significantly from the original source over time. This accumulation of errors makes it challenging to maintain visual coherence and fidelity, especially in long videos. A significant gap remains in addressing both global and local contexts, leading to inaccuracies and inconsistencies across the spatiotemporal dimension. 

\par
To address these challenges, we introduce \modelname{}, a unified spatiotemporal video adaptation framework designed for consistent and precise video editing, pushing the boundaries of editing minute-long videos, as shown in \cref{fig:teaser}. 
First, our framework introduces a novel \emph{test-time editing adaptation} mechanism that tune the image editing model on dataset generated by itself using the video to be edited, allowing the image editing model to learn associations between specific visual editing directions and corresponding instructions. This significantly enhances semantic comprehension and editing consistency within individual frames. To further improve local consistency, we introduce local latent adaptation to control local edits across frames, ensuring frame consistency before and after editing.

Second, effective editing requires seamless transitions and consistent edits, especially for long videos. To address this, we introduce \emph{spatiotemporal attention adaptation} to maintain global editing coherence across the edited frames. Specifically, we propose \gatherswap{} to \gath{} consistent attention variables from the model's architecture and strategically apply them throughout the video sequence. This approach not only aligns with the continuity of the video but also reinforces the fidelity of the editing process.
\par
Through rigorous evaluation, our methods have demonstrated superior performance compared to existing techniques, delivering significant improvements in both local edit precision and the overall aesthetic quality of the videos. Moreover, our approach is considerably faster than previous methods due to the parallelized swapping process. To the best of our knowledge, we are the first to achieve consistent editing of minute-long videos. Our main contributions are as follows: 
\begin{itemize}[leftmargin=*, itemsep=0.2em, labelsep=0.3em, topsep=0.2em, parsep=0em]
\item We introduce \modelname{}, a novel framework designed to enable \textbf{faithful, consistent, precise, and fast video editing.} Our approach pushes the boundaries of current video editing methods, ensuring both local and global consistency across the entire video.
\item We introduce a novel \textbf{spatiotemporal attention adaptation} and \textbf{test-time adaptation mechanism}, enabling coherent, text-driven video edits by maintaining global consistency across frames and semantic consistency within individual frames, leveraging an image editing model for video editing.
\item \textbf{Our approach outperforms existing techniques in human evaluation and automatic evaluation}, delivering significantly better performance in terms of editing quality and efficiency. 
\end{itemize}

\section{Related Work}
\label{sec:related_work}

\subsection{Text-driven Video Editing}
Text-driven video editing is a process of modifying videos according to the user's instructions. Inspired by the remarkable success of text-driven image editing~\citep{avrahami2022blended,brooks2023instructpix2pix,tumanyan2023plug,sheynin2023emu,Zhang2023MagicBrush}, extensive methods have been proposed for video content editing~\citep{ouyang2024codef,feng2024ccedit,li2024vidtome,yang2024fresco,zhangcontrolvideo,qin2023instructvid2vid,khachatryan2023text2video, tokenflow2023,wu2023fairy,qi2023fatezero,wang2023zero,ku2024anyv2v}. One paradigm for video editing is to adapt an image-based model to video. For example, \citet{khachatryan2023text2video} adapts image editing to the video domain without any training or fine-tuning by changing the self-attention mechanisms in Instruct-Pix2Pix to cross-frame attentions. \citet{tokenflow2023} explicitly propagates diffusion features based on inter-frame correspondences to enforce consistency in the diffusion feature space. \citet{yang2023neural} construct a neural video field to enable encoding long videos with hundreds of frames in a memory-efficient manner and then update the video field with an image-based model to impart text-driven editing effects. \citet{ku2024anyv2v} plug in any existing image editing tools to support an extensive array of video editing tasks. However, these methods are constrained by their ability to maintain global and local consistency, limiting to edit short videos within seconds. To efficiently enable longer video editing, \citet{wu2023fairy} centers on the concept of anchor-based cross-frame attention, firstly achieving editing 27-second videos. In our work, we built upon this line of work and improve editing consistency, firstly pushing the limits of editing to minutes-long videos.

\subsection{Spatiotemporal Consistency}
Ensuring spatiotemporal consistency is critical for video editing, especially for long videos.
\citet{qi2023fatezero} makes the attempt to study and utilize the cross-attention and spatial-temporal self-attention during DDIM inversion.
\citet{wang2023zero} proposes a spatial regularization module to fidelity to the original video.
\citet{park2024spectral} presents spectral motion alignment (SMA), a framework that learns motion patterns by incorporating frequency-domain regularization, facilitating the learning of whole-frame global motion dynamics, and mitigating spatial artifacts. \citet{ceylan2023pix2video} and \citet{wu2023tune} improve the design of spatial attention to cross-frame attention to ensure consistency. In our work, we further ensure consistency inside the anchor-based frames and propose a two-step gather-swap process to adapt spatiotemporal attention for consistent global editing.

\section{Preliminaries}
\label{sec:preliminary}
\noindent\textbf{Diffusion Models.}
In this work, we adapt an image editing model for instruction-based video editing. Given an image $x$, the diffusion process produces a noisy latent $\bs{z}_{t}$ from the encoded latent $z=\mathcal{E}(x)$ where the noise level increases over current timestep $t$ over total $T$ steps. A network $\bs{\epsilon}_{\theta}$ is trained to minimize the following optimization problem,
\begin{equation}
    \displaystyle\min_{\theta}\mathbb{E}_{y,\epsilon,t}\Big[ \big\lVert \epsilon - \epsilon_\theta(z_t, t, \mathcal{E}(c_I), c_T) \big\rVert \Big]
\end{equation}
where $\epsilon \in \mathcal{N}(0, 1)$ is the noise added by the diffusion process and $y = (c_T, c_I, x)$ is a triplet of instruction, input image and target image.
Here $\epsilon_\theta$ uses a U-Net architecture~\citep{ronneberger2015u}, including convolutional blocks, as well as self-attention and cross-attention layers.

\noindent\textbf{Attention Layer.}
The attention layer first computes the attention map using query, $\mathbf{Q} \in \mathbb{R}^{n_q \times d}$, and key, $\mathbf{K} \in \mathbb{R}^{n_k \times d}$ where $d$, $n_q$ and $n_k$ are the hidden dimension and the numbers of the query and key tokens respectively.  Then, the attention map is applied to the value, $\mathbf{V} \in \mathbb{R}^{n \times d}$ as follows:
\begin{gather}
    \mathbf{Z'}=\text{Attention}(\mathbf{Q},\mathbf{K},\mathbf{V}) = \text{Softmax}(\frac{\mathbf{Q}\mathbf{K}^{\top}}{\sqrt{d}})\mathbf{V},\\
    \label{eqn:attn}
    \mathbf{Q} = \mathbf{Z}\mathbf{W}_q, \;\; \mathbf{K} = \mathbf{C}\mathbf{W}_k,  \;\; \mathbf{V} = \mathbf{C}\mathbf{W}_v, 
\end{gather}
where $\mathbf{W}_q, \mathbf{W}_k, \mathbf{W}_v $ are the projection matrices to map the different inputs to the same hidden dimension $d$.  $\mathbf{Z}$ is the hidden state and $\mathbf{C}$ is the condition.  For self-attention layers, the condition is the hidden state, while the condition is text conditioning in cross-attention layers.

\noindent\textbf{Cross-frame Attention.}
Given $N$ frames from the source video, cross-frame attention has been employed in video editing by incorporating $\mathbf{K}$ and $\mathbf{V}$ from previous frames into the current frame's editing process~\citep{liu2023video, wang2023zero, wu2023fairy}, as shown below:
\begin{equation}
\label{eq:cross-frame-attention}
    \phi = \text{Softmax} \left( \frac{\mathbf{Q_\text{curr}} [\mathbf{K_\text{curr}}, \mathbf{K_\text{group}]^T}}{\sqrt{d}} \right) [\mathbf{V_\text{curr}}, \mathbf{V_\text{group}}],
\end{equation}
where  $\mathbf{K_\text{group}} = [\mathbf{K}^0, \dots, \mathbf{K}^k]$ and $\mathbf{V_\text{group}} = [\mathbf{V}^0, \dots, \mathbf{V}^k]$, and $k$ is the group size. By incorporating $\mathbf{K}_\text{group}$ and $\mathbf{V}_\text{group}$ during the video editing process for each frame, the temporal consistency is improved. In this paper, we improve cross-frame attention with a two-stage gather-swap process to significantly improve the spatiotemporal consistency.

\section{The \modelname{} Framework}

\begin{figure*}
    \centering
    \includegraphics[width=0.8\textwidth]{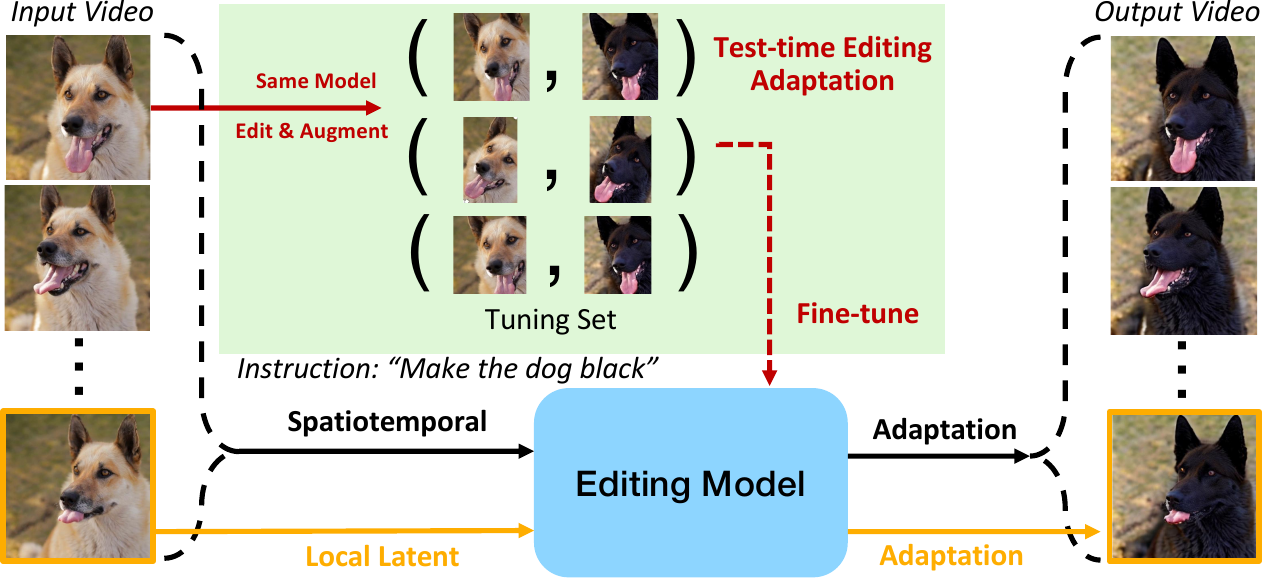}
    \caption{\textbf{Overview of \modelname{} framework.} For local consistency, Test-time Editing Adaptation finetunes the editing model with augmented editing pairs to ensure consistent editing directions with the text instruction, and Local Latent Adaptation achieves precise editing control and preserves non-target pixels from the input video. For global consistency, Spatiotemporal Adaptation collects and applies key attention variables across all frames.}
    \label{fig:model_overview}
    \vspace{-2.5ex}
\end{figure*}

Below, we outline the distinct methodologies that form the foundation of our approach.
We introduce a unified framework to tackle key challenges in instruction-guided video editing, with a focus on ensuring editing consistency and spatiotemporal coherence across video frames by leveraging an image editing model, as shown in \cref{fig:attention-process}. For a video to be edited, we first tune the editing direction of the editing model as the test-time adaptation in \cref{subsection:test-time-adaptation}, then edit each frame by Spatiotemporal Adaptation as in \cref{subsection:spatiotemporal}. With external masks, we could further achieve targeted editing.

\subsection{Test-Time Editing Adaptation for Local Consistency}
\label{subsection:test-time-adaptation}

When adapting image editing models for video editing, the same instructions must yield consistent semantic interpretations across frames—for example, every frame should exhibit the same degree of darkness when instructed to \emph{``make it night.''} Additionally, non-target elements in each frame must remain unchanged; for instance, a table should remain intact when the instruction is to replace an apple with an orange. To address these challenges, we propose two orthogonal approaches to achieve consistent local editing.

Inspired by DreamBooth~\citep{ruiz2023dream-booth}, which employs inference-time fine-tuning to associate specific objects with unique textual tokens, we similarly link visual editing outcomes with corresponding instructions, as shown in \cref{fig:model_overview}. We begin with a pipeline to generate the in-domain tuning set without the need for external resources. The image editing model $\Psi$ first edits a randomly sampled frame $S_\text{root}$ from the video to be edited to get editing result $E_\text{root}$. Then we apply random affine transformations to both the edited frame and source frame. Consider $\mathcal{F}_k$ as affine transformation:
\begin{equation}
    T = \{(\mathcal{F}_k(S), \mathcal{F}_k(E), I) \mid \mathcal{F}_k \in \mathcal{F}\}
\end{equation}
where $\mathcal{F}$ is the set of transformations. The tuning set $T$ consists of triples: source image, edited image, and editing instruction. Then the editing model is tuned on the triplets that is generated by itself from the video to be edited.
Therefore, the model learns to map specific visual editing directions to the corresponding instructions for the video. 

For the second challenge, where edits target specific areas, video models often unintentionally affect untargeted regions. In image editing, background preservation involves inverting the source image into latent space and blending it with the generated latent using a mask to control edits~\citep{cao2023masactrl, gu2024swapanything}. However, directly applying this approach to video editing causes severe glitching issues, as the generated areas do not stay aligned across frames. To address this, we propose \textbf{Local Latent Adaptation} in the context of video editing. The core behind it is \textbf{Progressive Boundary Integration}, which blends the inverted and generated latents at each timestep, confining edits to designated areas while preserving non-targeted regions. Please check Appendix for more details. Our approach ensures strict adherence to editing instructions, focusing solely on specified areas.
Our approach smoothly merges source and target latents via linear interpolation between 0 and 1 over the time series. The mathematical representation is given by:
\begin{equation}
\maskt{}(x, y) = 
\begin{cases}
\mask{}(x, y) \cdot \frac{t}{T}, & \text{if } t \leq T \text{ and } \mask{}(x, y) = 1 \\
\mask{}(x, y), & \text{otherwise}
\end{cases}
\end{equation}

\vspace{-3ex}

\begin{equation}
\label{equation:blending}
\bs{z}_{t}^{target} = \maskt{} \cdot \bs{z}_{t}^{edit} + (1 - \maskt{}) \cdot \bs{z}_{t}^{inverted}
\end{equation}

\vspace{-4ex}

\begin{equation}
\label{equation:sample}
\bs{z}_{t-1}^{edit} = Sample(\bs{z}_{t}^{target}, \Phi, t)
\end{equation}

Here, $\mask{}$ is the giving binary mask and \( \mask{}(x, y) \) is predefined as 1 in a target area and 0 elsewhere. Within this central area, \( \mask{}(x, y) \) incrementally decrease from 0 to 1 over \( T \) steps, while the values outside this central region remain unchanged.  By applying external masks to define the editing region as in \cref{equation:blending} and then sample the latent for the next diffusion step as in \cref{equation:sample} iteratively, VIA was able to achieve targeted editing. Note that other parameters such as editing instruction are ignored for simplicity. To assist \modelname{} framework, we built a mask generation process as in the Appendix.

\subsection{Spatiotemporal Adaptation for Global Consistency}
\label{subsection:spatiotemporal}
\vspace{-1ex}
\begin{figure*}
    \centering
    \includegraphics[width=0.8\textwidth]{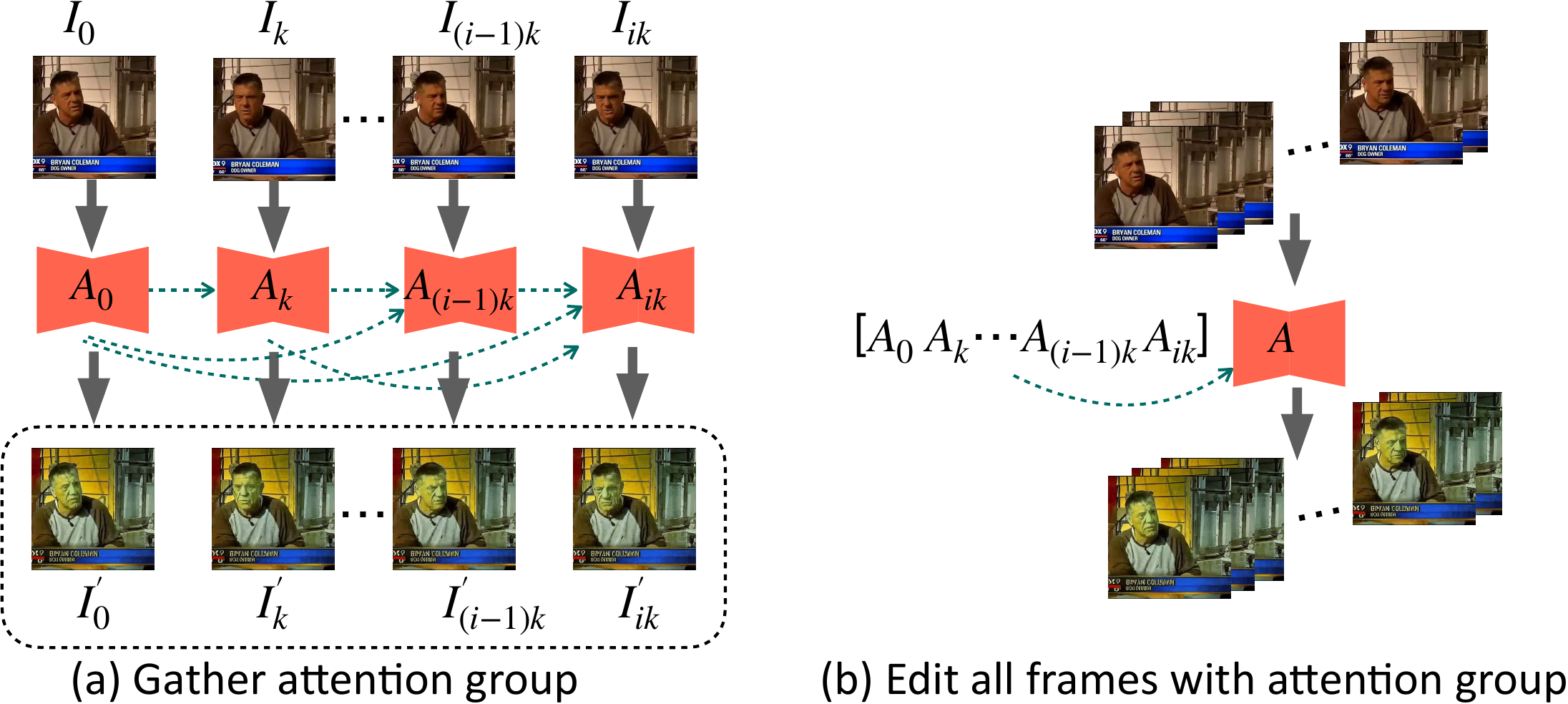}
    \caption{\textbf{The \gatherswap{} process for video editing.} The left part of the diagram illustrates the gathering process. We initially sample $k+1$ frames evenly distributed throughout the video. The first frame undergoes standard editing using an image editing model, during which the attention variables are captured and stored. For each of the subsequent $k$ frames, the attention variable from the preceding frame is swapped in, and its own attention variables are also preserved. In the right part, the collected attention variables from all $k+1$ frames are swapped into the editing process of each frame. This includes applying the previously gathered attention variables to enhance the consistency and quality of edits across the sequence.}
    \label{fig:attention-process}
    \vspace{-2ex}
\end{figure*}

For long video editing, maintaining smooth transitions without glitches or artifacts is essential. Attention variables within the U-net have been found to correlate strongly with the generated content. To ensure consistent global editing, we propose a two-step \gatherswap{} process to adapt spatiotemporal attention, as illustrated in \cref{fig:attention-process}. In this method, the gathered group is uniformly applied across all frames, ensuring internal coherence throughout the editing process.

Firstly, in the \gath{} stage, the model progressively edits the image, with key $\mathbf{K}$ and value $\mathbf{V}$ from previous frames in the group, rather from their own $\mathbf{K}_\text{curr}$ and $\mathbf{V}_\text{curr}$,

\begin{equation}
    \phi = \softmax \left( \frac{\mathbf{Q}_\text{curr}\mathbf{K}_\text{prev}^T}{\sqrt{d}} \right) \mathbf{V}_\text{prev},
\end{equation}

\vspace{-4ex}

\begin{equation}
    \mathbf{K}_\text{group}^{(t+1)} = [\mathbf{K}_\text{group}^{(t)}, \mathbf{K}_\text{curr}], \quad \mathbf{V}_\text{group}^{(t+1)} = [\mathbf{V}_\text{group}^{(t)}, \mathbf{V}_\text{curr}]
\end{equation}
Since $\mathbf{K}_\text{curr}$ and $\mathbf{V}_\text{curr}$ are calculated by the $\phi$ from the last layer, which already has a stronger dependency on other frames, the saved elements have a stronger consistency with previous group elements, leading to in-group consistency in $\mathbf{K}_\text{group}^{(k+1)}$ and $\mathbf{V}_\text{group}^{(k+1)}$.

In the second stage, we apply the attention group to the editing process of all frames, including those used initially to generate the attention group. Expanding $K$ and $V$ does not change the output, as $QK^T$ remains structured, and multiplication with $V$ keeps the dependency on $Q$ and $V$. Thus, a signal can integrate information from multiple others. This approach resolves the inconsistency in the group frames, where they initially have less dependency on other frames. Throughout the editing process, each frame continues to refrain from using its own attention variables, instead relying on the shared attention group to maintain consistency across the entire video. This ensures that all frames, even the earlier ones, are edited with a global perspective, reducing discrepancies between frames.
\begin{equation}
    \phi = \softmax \left( \frac{\mathbf{Q}_\text{curr}\mathbf{K}_\text{group}^T}{\sqrt{d}} \right) \mathbf{V}_\text{group},
\end{equation}
In this way, all frames share the same attention group, leading to maximum coherence between the edited frames and enabling the \swap{} process to be distributed across multiple GPUs, which significantly reduces editing time.
Moreover, while previous work has primarily relied on self-attention for cross-frame consistency, we discovered that cross-attention also plays a crucial role in maintaining coherence. Combining both self-attention and cross-attention mechanisms capturing a broad representation of frame differences and maximizing consistency in the edits. \cref{fig:attention-process} illustrates the two stages, where $\mathbf{A}$ represents both $\mathbf{K}$ and $\mathbf{V}$.

\begin{table*}[hbt]
\centering
    \caption{\textbf{Human evaluation results.} We compare our model with five previous open-source methods from three aspects. `Tie' indicates the two models are on par with each other. Only spatiotemporal adaptation is used when compared with baseline models. }
    \vspace{-2ex}
    \setlength{\tabcolsep}{2.0pt}
    \resizebox{\linewidth}{!}{
    \begin{tabular}{c|ccc|ccc|ccc|ccc|ccc}
        \toprule
        ~ & Ours & Rerender & Tie & Ours & TokenFlow & Tie & Ours & AnyV2V & Tie  & Ours & Video-P2P & Tie &
        Ours & Tune-A-Video & Tie\\
        \midrule
        Instruction Following ~ & \textbf{50.50} & 34.00  & 15.5 & \textbf{75.75} & 16.00 & 8.25  & \textbf{56.00} & 29.00 & 15.00 & \textbf{74.00} & 16.25 & 9.75 & \textbf{70.25} & 20.75 & 9.00\\
        Consistency ~ & \textbf{47.25} & 35.00 & 17.75 & \textbf{38.00} & 31.50 & 30.5 & \textbf{53.50} & 23.25 & 23.25 & \textbf{80.50} & 9.50 & 10.00 & \textbf{68.75} & 20.75 & 10.5\\
        Overall Quality ~ & \textbf{53.50}  & 29.00 & 17.5 & \textbf{61.75} & 22.75 & 15.5 & \textbf{63.50} & 30.00 & 6.5 & \textbf{63.75} & 22.75 & 13.5& \textbf{56.00} & 22.25 & 21.75 \\
        \bottomrule
    \end{tabular}
    }
    \label{table:human-evaluation}
\end{table*}

\begin{table*}[hbt]
    \caption{\textbf{Automatic evaluation results.} \modelname{} outperforms open-sourced video editing models in automatic metrics. Only spatiotemporal adaptation is used when compared with baseline models.}
    \vspace{-2ex}
\centering
\resizebox{0.9\linewidth}{!}{
    \begin{tabular}{ccccccc}
        \toprule
        ~ & \modelname{} & Rerender & TokenFlow & AnyV2V &  Video-P2P & Tune-A-Video \\
        \midrule
        Frame-Acc $\uparrow$ ~ & \textbf{0.869} & 0.734 & 0.587 & 0.533 & 0.587 & 0.601\\
        Tem-Con $\uparrow$ ~ & \textbf{0.983} & 0.954 & 0.932  & 0.856 & 0.912 & 0.927 \\
        Pixel-MSE $\downarrow$ ~ & \textbf{0.011} & 0.016  & 0.018  & 0.026 & 0.020 & 0.019  \\
        Latency(sec) $\downarrow$  & \textbf{16} & 406 & 450 & 570 & 612 & 529 \\
        \bottomrule
    \end{tabular}
    \vspace{-2ex}
    }
    \label{table:automatic-evaluation}
\end{table*}

\section{Evaluation}

In this paper, we adapt image editing model MGIE~\citep{fu2023guiding} for video editing. Please refer to the Appendix for performance on other backbone. We conduct both qualitative and human evaluations against open-source state-of-the-art baselines, including Fairy~\citep{wu2023fairy}, AnyV2V~\citep{ku2024anyv2v}, Rerender~\citep{yang2023rerender}, Tokenflow~\citep{tokenflow2023}, Video-P2P~\citep{liu2023video}, and Tune-A-Video~\citep{wu2023tune}. For the comparison with AnyV2V, we use the first edited frame generated by \modelname{} as the starting point for the evaluation. Please refer to the Appendix for details about the implementation process of the baselines. We used 800 videos for the test set, where 400 of them are short video, and the remaining range from 1 minutes to 2 minutes. Short videos are collected from Panda-70M and long videos are from https://www.shutterstock.com/video.

\subsection{Quantitative Evaluation}
\noindent\textbf{Human Evaluation.~}
We began by conducting a human evaluation. Since many baselines are unable to handle long videos, we limited the video length to 4–8 seconds to ensure a fair comparison. All videos were standardized to a frame size of 512x512 pixels. A total of 400 videos were sampled for human evaluation to compare the performance.
The evaluation focused on three key criteria: \textbf{Instruction Following}, assessing accuracy in executing user commands; \textbf{Consistency}, ensuring coherence across frames without abrupt changes; and \textbf{Overall Quality}, gauging visual appeal and smoothness. Results in \cref{table:human-evaluation} show that \modelname{} excelled in all metrics compared with other baselines.

\begin{figure*}[bht!]
    \centering
    \includegraphics[width=1.0\textwidth]{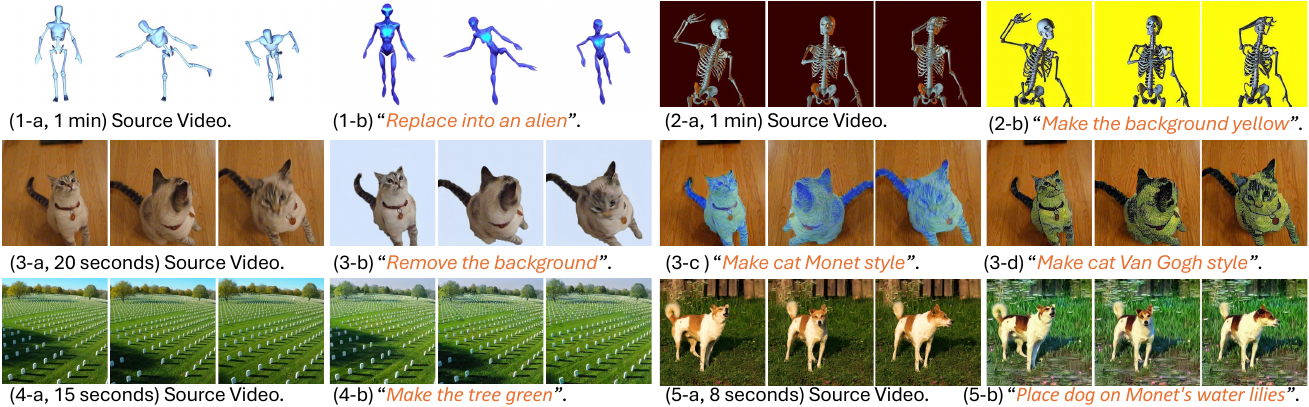}
    \vspace{-3ex}
    \caption{\textbf{Local editing results.} \modelname{} is capable of performing a wide range of localized editing tasks, where only specific regions or pixels within a frame are modified. The video length is introduced in the text below the video frames.}
    \label{fig:local-result}
\end{figure*}

\begin{figure*}[htb!]
    \centering
    \includegraphics[width=1.0\textwidth]{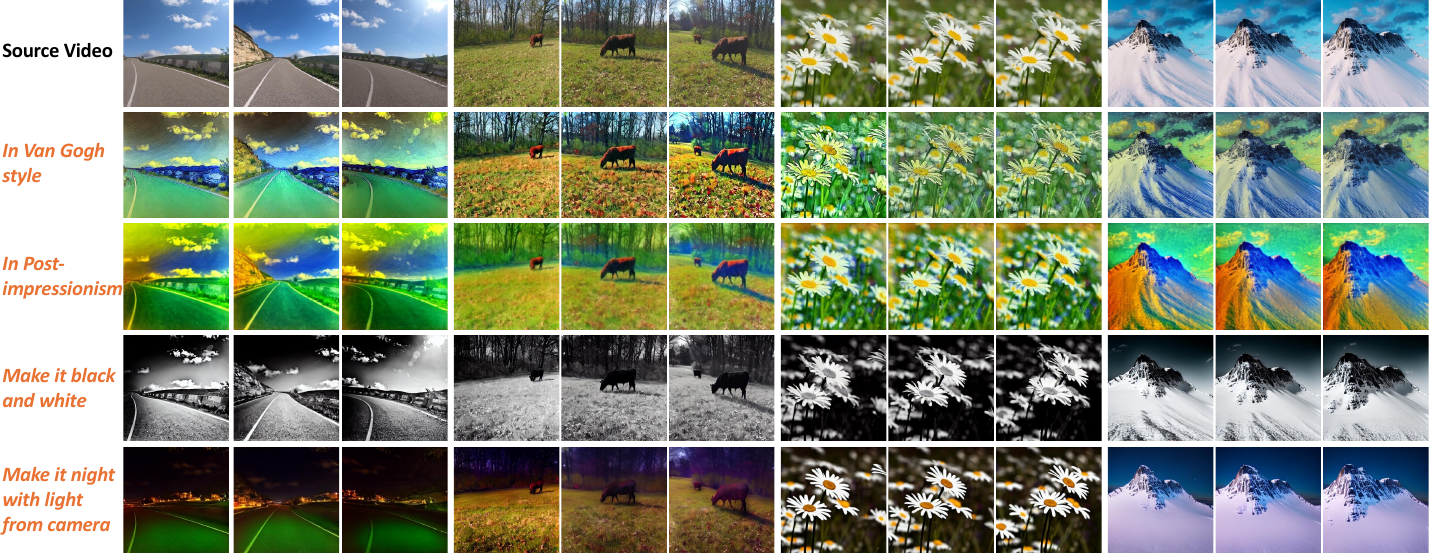}
    \vspace{-3ex}
    \caption{\textbf{Global editing results.} \modelname{} demonstrates robust global editing performance across various videos using a consistent set of editing instructions, producing high-quality results. The videos are of length 2-minute, 1-minute video, 30 seconds, and 7 seconds.}
    \label{fig:global-result}
    \vspace{-1ex}
\end{figure*}

\noindent\textbf{Automatic Evaluation.~}
We also conducted automatic evaluation as in \cref{table:automatic-evaluation}. Frame-Acc~\citep{qi2023fatezero, yang2023rerender} measures the percentage of
frames where the edited image has a higher CLIP similarity
to the target prompt than the source prompt; Tem-Con~\citep{esser2023structure} measures the temporal consistency via computing the cosine similarity between all pairs of consecutive frames. Following~\cite{ceylan2023pix2video}, we also use Pixel-MSE to calculate the difference between the edited frame and its previous frame warped with the optical flow calculated from the source frame pairs. Note that it is normalized by the maximum possible MSE difference. \modelname{} outperformed all other models across these metrics, delivering superior accuracy and consistency while also achieving faster processing speeds. We did not use test-time adaptation for \modelname{}, as some of the baseline models do not inherently benefit from it, which ensured a fair comparison.
Additionally, we calculated the evaluation latency of the editing process, which was carried out on an A100 machine with 8 GPUs. The global adaptation process could be distributed across multiple GPUs to further accelerate the process. Detailed speed analysis can be found in the Appendix.

\subsection{Qualitative Results}
\noindent\textbf{Local Editing Results.} \cref{fig:local-result} showcases the performance of \modelname{} on various local editing tasks, where only specific parts of the frame are modified. \modelname{} excels at accurately identifying the target area and applying precise edits. \modelname{} demonstrate strong performance on general local editing tasks including both \textbf{background modification} and \textbf{foreground object modification}. The two 1-min long video in the first row speficially presented its precise control. Besides, \modelname{} enables local stylization, surpassing traditional techniques limited to full-image changes, whose enhanced control opens up new creative possibilities in video editing.

\noindent\textbf{Global Editing Results.} \cref{fig:global-result} highlights the global editing capabilities of \modelname{} across a range of videos. A uniform set of editing instructions was used across different videos, resulting in coherent and visually appealing modifications throughout. The bottom example specifically illustrates \modelname{}'s proficiency in understanding and consistently applying visual effects across all frames, ensuring seamless transitions and maintaining the integrity of the visual narrative across the entire video.

\begin{figure*}[htb!]
    \centering
    \includegraphics[width=1.0\textwidth]{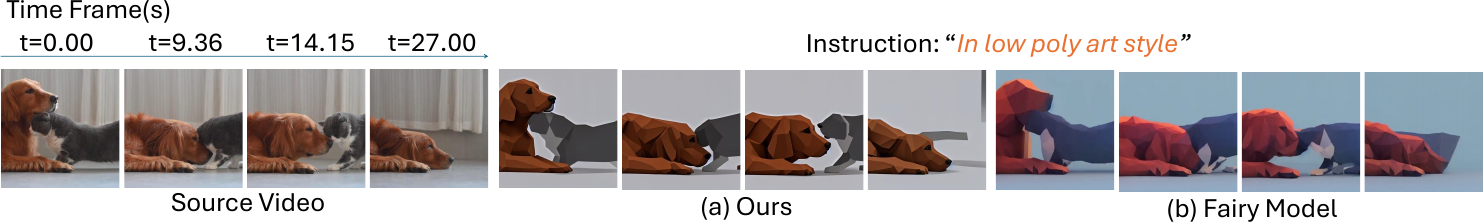}
    \vspace{-3ex}
    \caption{\textbf{Comparison with the baseline model on the long video.} We present the editing results from a 27-second video.}
    \label{fig:comp-fairy}
\end{figure*}

\noindent\textbf{Long Video Editing.} 
A direct consequence of the high consistency feature in our video editing framework is its proficiency in handling longer videos, as demonstrated throughout this paper. Currently, existing video editing models cannot handle minute-long videos due to architectural limitations, making direct comparisons challenging. To address this, we evaluate long video editing by concatenating individually edited chunks, where \modelname{} significantly outperforms the baselines. For more details, see \cref{appendix:long-video-comparison}. One of our baselines, Fairy~\citep{wu2023fairy}, has not made their code publicly available, but they report that their model supports videos up to 27 seconds in length. We compare our results on the same video in their website using identical editing instructions, as shown in \cref{fig:comp-fairy}. \modelname{} demonstrates superior global and local consistency, which can be attributed to our unified adaptation framework.%

\vspace{-0ex}

\begin{figure*}
    \centering
    \includegraphics[width=1.0\textwidth]{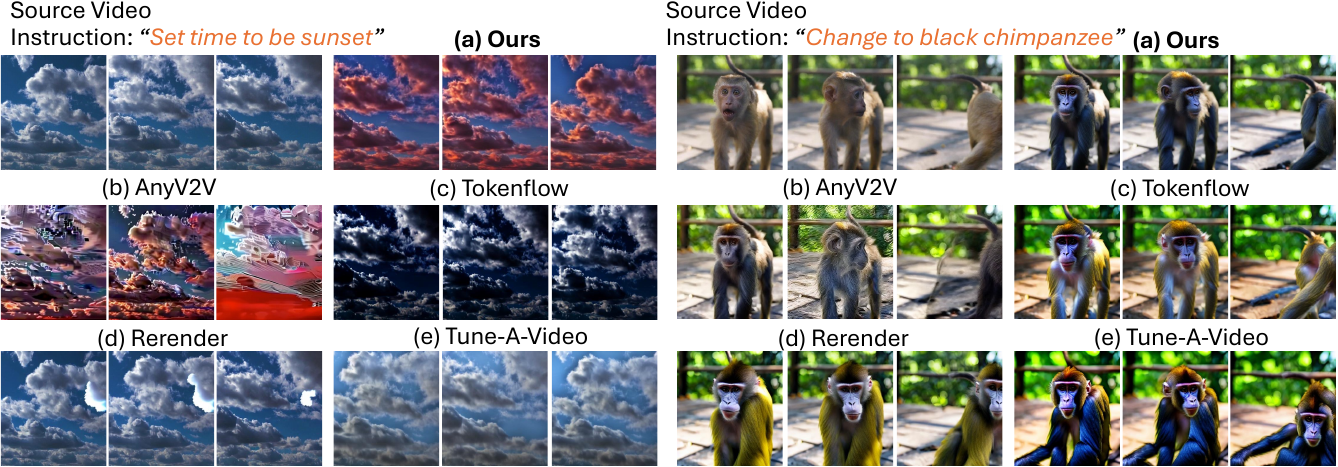}
    \vspace{-3ex}
    \caption{\textbf{Qualitative comparison with baselines.} \modelname{} is able to produce consistent editing results.}
    \label{fig:qualitative-comparison}
    \vspace{-1ex}
\end{figure*}

\begin{figure*}[htb!]
    \centering
    \includegraphics[width=1.0\textwidth]{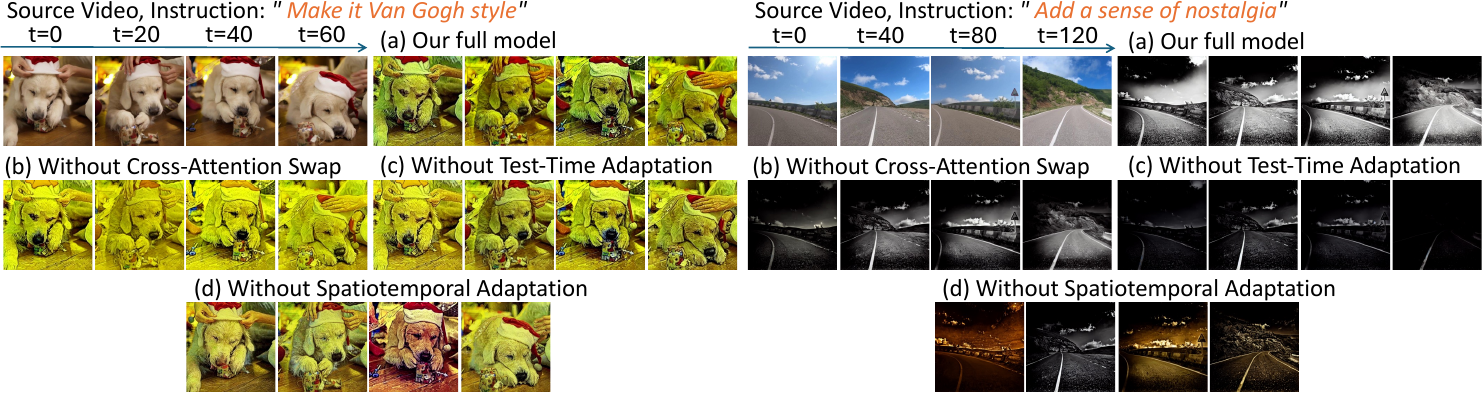}
    \caption{\textbf{Ablation Study on components in \modelname{} on long video.} In the left example, the hat color and visual style are less consistent without distinct component handling. In contrast, the right example shows a uniform visual style applied consistently across frames, with each component maintaining its appearance. Test-time adaptation ensures stable visual effects that follow the specified instructions. Without the gather-swap technique, object consistency across frames is weakened. Additionally, incorporating cross-attention alongside self-attention improves consistency and reduces artifacts.
    }
    \vspace{-1.5em}
    \label{fig:ablation-long}
\end{figure*}

\noindent\textbf{Qualitative Comparison.} In \cref{fig:qualitative-comparison}, we present two examples of video editing to showcase the performance of \modelname{} in comparison to other models. 
In the first example, the video depicts rapidly moving clouds against a blue sky, with the instruction to "Set the time to sunset." Despite the swift movement of the clouds, which places a high demand on temporal consistency, \modelname{} demonstrates excellent coherence across frames. The Editing Adaptation process allows \modelname{} to effectively align the visual effect with the concept of "sunset," ensuring smooth and realistic changes. In contrast, other models struggled to execute the command adequately. The AnyV2V model partially achieved the desired visual effect by leveraging the initial frame generated by \modelname{}.
On the right, we show an object-swapping example where a monkey moves from within the frame to outside of it. The challenge lies in maintaining a smooth transition from the full subject to a partially visible one. While other methods often introduce artifacts between the edited frames and the original video, \modelname{} seamlessly swaps the subject's identity, preserving visual coherence and continuity throughout the transition.

From this comparison, we found that (1) \modelname{} outperforms the baselines in both editing quality and processing speed. It ensures smooth transitions in edited videos, even when dealing with rapidly moving objects, while some models, such as AnyV2V, generate noticeable artifacts. (2) \modelname{} demonstrates strong performance in adhering to complex instructions, where other models often struggle. While competing methods experience degraded performance with intricate commands, \modelname{} consistently follows the instructions, applying edits accurately across all frames.

\noindent\textbf{Ablation on Individual Components.} In \cref{fig:ablation-long}, we analyze the impact of various components of \modelname{} on the editing of long videos. Our experiments indicate that the quality of the initial edited frames plays a critical role in determining the overall visual quality, as information from these root frames propagates throughout the video sequence. Test-time adaptation further enhances the model’s ability to closely follow the editing instructions, improving overall consistency.
When \gatherswap{} is omitted and the model relies solely on cross-frame attention, inconsistencies start to emerge between frames. Additionally, although self-attention is commonly employed to ensure consistency, we found that the inclusion of cross-attention significantly improves the quality of video editing. In the left example, the hat color and visual style lack consistency due to the absence of distinct component handling. In contrast, the right example demonstrates a cohesive visual style applied uniformly across frames, with each component retaining its appearance. For additional ablation studies, and analysis on detailed components such as Progressive Boundary Integration, please refer to the Appendix.

\vspace{-2ex}

\section{Conclusion}

This paper introduces a novel video editing framework that tackles the critical challenges of achieving temporal consistency and precise local edits. Our approach surpasses the limitations of traditional frame-by-frame methods, delivering coherent and immersive video experiences. Extensive experiments show that our framework outperforms existing baselines in terms of handling temporal dynamics, ensuring local edit precision, and enhancing overall video aesthetic quality. This advancement paves the way for new possibilities in media production and creative content generation, setting a new benchmark for future developments in video editing technology.

\clearpage

{
    \small
    \bibliographystyle{ieeenat_fullname}
    \bibliography{main}

\begin{thebibliography}{47}
\providecommand{\natexlab}[1]{#1}
\providecommand{\url}[1]{\texttt{#1}}
\expandafter\ifx\csname urlstyle\endcsname\relax
  \providecommand{\doi}[1]{doi: #1}\else
  \providecommand{\doi}{doi: \begingroup \urlstyle{rm}\Url}\fi

\bibitem[Avrahami et~al.(2022)Avrahami, Lischinski, and Fried]{avrahami2022blended}
Omri Avrahami, Dani Lischinski, and Ohad Fried.
\newblock Blended diffusion for text-driven editing of natural images.
\newblock In \emph{CVPR}, 2022.

\bibitem[Brooks et~al.(2023)Brooks, Holynski, and Efros]{brooks2023instructpix2pix}
Tim Brooks, Aleksander Holynski, and Alexei~A Efros.
\newblock Instructpix2pix: Learning to follow image editing instructions.
\newblock In \emph{CVPR}, pages 18392--18402, 2023.

\bibitem[Calandra et~al.(2008)Calandra, Gurvitch, and Lund]{calandra2008exploratory}
Brendan Calandra, Rachel Gurvitch, and Jacalyn Lund.
\newblock An exploratory study of digital video editing as a tool for teacher preparation.
\newblock \emph{Journal of Technology and Teacher Education}, 16\penalty0 (2):\penalty0 137--153, 2008.

\bibitem[Calandra et~al.(2009)Calandra, Brantley-Dias, Lee, and Fox]{calandra2009using}
Brendan Calandra, Laurie Brantley-Dias, John~K Lee, and Dana~L Fox.
\newblock Using video editing to cultivate novice teachers’ practice.
\newblock \emph{Journal of research on technology in education}, 42\penalty0 (1):\penalty0 73--94, 2009.

\bibitem[Cao et~al.(2023)Cao, Wang, Qi, Shan, Qie, and Zheng]{cao2023masactrl}
Mingdeng Cao, Xintao Wang, Zhongang Qi, Ying Shan, Xiaohu Qie, and Yinqiang Zheng.
\newblock Masactrl: Tuning-free mutual self-attention control for consistent image synthesis and editing.
\newblock \emph{arXiv preprint arXiv:2304.08465}, 2023.

\bibitem[Ceylan et~al.(2023)Ceylan, Huang, and Mitra]{ceylan2023pix2video}
Duygu Ceylan, Chun-Hao~P Huang, and Niloy~J Mitra.
\newblock Pix2video: Video editing using image diffusion.
\newblock In \emph{Proceedings of the IEEE/CVF International Conference on Computer Vision}, pages 23206--23217, 2023.

\bibitem[Chen et~al.(2024)Chen, Siarohin, Menapace, Deyneka, Chao, Jeon, Fang, Lee, Ren, Yang, and Tulyakov]{chen2024panda70m}
Tsai-Shien Chen, Aliaksandr Siarohin, Willi Menapace, Ekaterina Deyneka, Hsiang-wei Chao, Byung~Eun Jeon, Yuwei Fang, Hsin-Ying Lee, Jian Ren, Ming-Hsuan Yang, and Sergey Tulyakov.
\newblock Panda-70m: Captioning 70m videos with multiple cross-modality teachers.
\newblock \emph{arXiv preprint arXiv:2402.19479}, 2024.

\bibitem[Dancyger(2018)]{dancyger2018technique}
Ken Dancyger.
\newblock \emph{The technique of film and video editing: history, theory, and practice}.
\newblock Routledge, 2018.

\bibitem[Esser et~al.(2023)Esser, Chiu, Atighehchian, Granskog, and Germanidis]{esser2023structure}
Patrick Esser, Johnathan Chiu, Parmida Atighehchian, Jonathan Granskog, and Anastasis Germanidis.
\newblock Structure and content-guided video synthesis with diffusion models.
\newblock In \emph{ICCV}, 2023.

\bibitem[Feng et~al.(2024)Feng, Weng, Wang, Yuan, Bao, Luo, Chen, and Guo]{feng2024ccedit}
Ruoyu Feng, Wenming Weng, Yanhui Wang, Yuhui Yuan, Jianmin Bao, Chong Luo, Zhibo Chen, and Baining Guo.
\newblock Ccedit: Creative and controllable video editing via diffusion models.
\newblock In \emph{Proceedings of the IEEE/CVF Conference on Computer Vision and Pattern Recognition}, pages 6712--6722, 2024.

\bibitem[Frierson(2018)]{frierson2018film}
Michael Frierson.
\newblock \emph{Film and Video Editing Theory}.
\newblock Routledge, 2018.

\bibitem[Fu et~al.(2024)Fu, Hu, Du, Wang, Yang, and Gan]{fu2023guiding}
Tsu-Jui Fu, Wenze Hu, Xianzhi Du, William~Yang Wang, Yinfei Yang, and Zhe Gan.
\newblock Guiding instruction-based image editing via multimodal large language models.
\newblock In \emph{ICLR}, 2024.

\bibitem[Geyer et~al.(2024)Geyer, Bar-Tal, Bagon, and Dekel]{tokenflow2023}
Michal Geyer, Omer Bar-Tal, Shai Bagon, and Tali Dekel.
\newblock Tokenflow: Consistent diffusion features for consistent video editing.
\newblock \emph{ICLR}, 2024.

\bibitem[Gu et~al.(2023)Gu, Wang, Zhao, Fu, Xiong, Liu, Zhang, Zhang, Zhang, Jung, and Wang]{gu2023photoswap}
Jing Gu, Yilin Wang, Nanxuan Zhao, Tsu-Jui Fu, Wei Xiong, Qing Liu, Zhifei Zhang, He Zhang, Jianming Zhang, HyunJoon Jung, and Xin~Eric Wang.
\newblock Photoswap: Personalized subject swapping in images, 2023.

\bibitem[Gu et~al.(2024)Gu, Wang, Zhao, Xiong, Liu, Zhang, Zhang, Zhang, Jung, and Wang]{gu2024swapanything}
Jing Gu, Yilin Wang, Nanxuan Zhao, Wei Xiong, Qing Liu, Zhifei Zhang, He Zhang, Jianming Zhang, HyunJoon Jung, and Xin~Eric Wang.
\newblock Swapanything: Enabling arbitrary object swapping in personalized visual editing.
\newblock \emph{arXiv preprint arXiv:2404.05717}, 2024.

\bibitem[Guo et~al.(2023)Guo, Yang, Rao, Liang, Wang, Qiao, Agrawala, Lin, and Dai]{guo2023animatediff}
Yuwei Guo, Ceyuan Yang, Anyi Rao, Zhengyang Liang, Yaohui Wang, Yu Qiao, Maneesh Agrawala, Dahua Lin, and Bo Dai.
\newblock Animatediff: Animate your personalized text-to-image diffusion models without specific tuning, 2023.

\bibitem[Hertz et~al.(2022)Hertz, Mokady, Tenenbaum, Aberman, Pritch, and Cohen-or]{hertz2022prompt}
Amir Hertz, Ron Mokady, Jay Tenenbaum, Kfir Aberman, Yael Pritch, and Daniel Cohen-or.
\newblock Prompt-to-prompt image editing with cross-attention control.
\newblock In \emph{The Eleventh International Conference on Learning Representations}, 2022.

\bibitem[Ho and Salimans(2022)]{ho2022classifier}
Jonathan Ho and Tim Salimans.
\newblock Classifier-free diffusion guidance.
\newblock \emph{arXiv preprint arXiv:2207.12598}, 2022.

\bibitem[Jackson(2016)]{jackson2016digital}
Wallace Jackson.
\newblock \emph{Digital video editing fundamentals}.
\newblock Springer, 2016.

\bibitem[Khachatryan et~al.(2023)Khachatryan, Movsisyan, Tadevosyan, Henschel, Wang, Navasardyan, and Shi]{khachatryan2023text2video}
Levon Khachatryan, Andranik Movsisyan, Vahram Tadevosyan, Roberto Henschel, Zhangyang Wang, Shant Navasardyan, and Humphrey Shi.
\newblock Text2video-zero: Text-to-image diffusion models are zero-shot video generators.
\newblock In \emph{Proceedings of the IEEE/CVF International Conference on Computer Vision}, pages 15954--15964, 2023.

\bibitem[Kholisoh et~al.(2021)Kholisoh, Andika, and Suhendra]{kholisoh2021short}
Nur Kholisoh, Dicky Andika, and Suhendra Suhendra.
\newblock Short film advertising creative strategy in postmodern era within software video editing.
\newblock \emph{Bricolage: Jurnal Magister Ilmu Komunikasi}, 7\penalty0 (1):\penalty0 041--058, 2021.

\bibitem[Kirillov et~al.(2023)Kirillov, Mintun, Ravi, Mao, Rolland, Gustafson, Xiao, Whitehead, Berg, Lo, Dollar, and Girshick]{Kirillov_2023_ICCV}
Alexander Kirillov, Eric Mintun, Nikhila Ravi, Hanzi Mao, Chloe Rolland, Laura Gustafson, Tete Xiao, Spencer Whitehead, Alexander~C. Berg, Wan-Yen Lo, Piotr Dollar, and Ross Girshick.
\newblock Segment anything.
\newblock In \emph{Proceedings of the IEEE/CVF International Conference on Computer Vision (ICCV)}, pages 4015--4026, 2023.

\bibitem[Ku et~al.(2024)Ku, Wei, Ren, Yang, and Chen]{ku2024anyv2v}
Max Ku, Cong Wei, Weiming Ren, Huan Yang, and Wenhu Chen.
\newblock Anyv2v: A plug-and-play framework for any video-to-video editing tasks.
\newblock \emph{arXiv preprint arXiv:2403.14468}, 2024.

\bibitem[Li et~al.(2024)Li, Ma, Yang, and Yang]{li2024vidtome}
Xirui Li, Chao Ma, Xiaokang Yang, and Ming-Hsuan Yang.
\newblock Vidtome: Video token merging for zero-shot video editing.
\newblock In \emph{Proceedings of the IEEE/CVF Conference on Computer Vision and Pattern Recognition}, pages 7486--7495, 2024.

\bibitem[Liu et~al.(2023{\natexlab{a}})Liu, Li, Wu, and Lee]{liu2024visual}
Haotian Liu, Chunyuan Li, Qingyang Wu, and Yong~Jae Lee.
\newblock Visual instruction tuning.
\newblock In \emph{NeurIPS}, 2023{\natexlab{a}}.

\bibitem[Liu et~al.(2023{\natexlab{b}})Liu, Zhang, Li, Lin, and Jia]{liu2023video}
Shaoteng Liu, Yuechen Zhang, Wenbo Li, Zhe Lin, and Jiaya Jia.
\newblock Video-p2p: Video editing with cross-attention control.
\newblock \emph{arXiv preprint arXiv:2303.04761}, 2023{\natexlab{b}}.

\bibitem[Mei et~al.(2007)Mei, Hua, Yang, and Li]{mei2007videosense}
Tao Mei, Xian-Sheng Hua, Linjun Yang, and Shipeng Li.
\newblock Videosense: towards effective online video advertising.
\newblock In \emph{Proceedings of the 15th ACM international conference on Multimedia}, pages 1075--1084, 2007.

\bibitem[Nichol et~al.(2022)Nichol, Dhariwal, Ramesh, Shyam, Mishkin, Mcgrew, Sutskever, and Chen]{nichol2022glide}
Alexander~Quinn Nichol, Prafulla Dhariwal, Aditya Ramesh, Pranav Shyam, Pamela Mishkin, Bob Mcgrew, Ilya Sutskever, and Mark Chen.
\newblock Glide: Towards photorealistic image generation and editing with text-guided diffusion models.
\newblock In \emph{ICML}, pages 16784--16804, 2022.

\bibitem[Ouyang et~al.(2024)Ouyang, Wang, Xiao, Bai, Zhang, Zheng, Zhou, Chen, and Shen]{ouyang2024codef}
Hao Ouyang, Qiuyu Wang, Yuxi Xiao, Qingyan Bai, Juntao Zhang, Kecheng Zheng, Xiaowei Zhou, Qifeng Chen, and Yujun Shen.
\newblock Codef: Content deformation fields for temporally consistent video processing.
\newblock In \emph{Proceedings of the IEEE/CVF Conference on Computer Vision and Pattern Recognition}, pages 8089--8099, 2024.

\bibitem[Park et~al.(2024)Park, Jeong, Lee, and Ye]{park2024spectral}
Geon~Yeong Park, Hyeonho Jeong, Sang~Wan Lee, and Jong~Chul Ye.
\newblock Spectral motion alignment for video motion transfer using diffusion models.
\newblock \emph{arXiv preprint arXiv:2403.15249}, 2024.

\bibitem[Podell et~al.(2023)Podell, English, Lacey, Blattmann, Dockhorn, M{\"u}ller, Penna, and Rombach]{podell2023sdxl}
Dustin Podell, Zion English, Kyle Lacey, Andreas Blattmann, Tim Dockhorn, Jonas M{\"u}ller, Joe Penna, and Robin Rombach.
\newblock Sdxl: Improving latent diffusion models for high-resolution image synthesis.
\newblock \emph{arXiv preprint arXiv:2307.01952}, 2023.

\bibitem[Qi et~al.(2023)Qi, Cun, Zhang, Lei, Wang, Shan, and Chen]{qi2023fatezero}
Chenyang Qi, Xiaodong Cun, Yong Zhang, Chenyang Lei, Xintao Wang, Ying Shan, and Qifeng Chen.
\newblock Fatezero: Fusing attentions for zero-shot text-based video editing.
\newblock In \emph{Proceedings of the IEEE/CVF International Conference on Computer Vision}, pages 15932--15942, 2023.

\bibitem[Qin et~al.(2023)Qin, Li, Tang, Chua, and Zhuang]{qin2023instructvid2vid}
Bosheng Qin, Juncheng Li, Siliang Tang, Tat-Seng Chua, and Yueting Zhuang.
\newblock Instructvid2vid: Controllable video editing with natural language instructions.
\newblock \emph{arXiv preprint arXiv:2305.12328}, 2023.

\bibitem[Ronneberger et~al.(2015)Ronneberger, Fischer, and Brox]{ronneberger2015u}
Olaf Ronneberger, Philipp Fischer, and Thomas Brox.
\newblock U-net: Convolutional networks for biomedical image segmentation.
\newblock In \emph{MICCAI}. Springer, 2015.

\bibitem[Ruiz et~al.(2023)Ruiz, Li, Jampani, Pritch, Rubinstein, and Aberman]{ruiz2023dream-booth}
Nataniel Ruiz, Yuanzhen Li, Varun Jampani, Yael Pritch, Michael Rubinstein, and Kfir Aberman.
\newblock {DreamBooth: Fine Tuning Text-to-Image Diffusion Models for Subject-Driven Generation}.
\newblock In \emph{CVPR}, 2023.

\bibitem[Schmitz et~al.(2006)Schmitz, Shafton, Shaw, Tripodi, Williams, and Yang]{schmitz2006international}
Patrick Schmitz, Peter Shafton, Ryan Shaw, Samantha Tripodi, Brian Williams, and Jeannie Yang.
\newblock International remix: video editing for the web.
\newblock In \emph{Proceedings of the 14th ACM international conference on Multimedia}, pages 797--798, 2006.

\bibitem[Sheynin et~al.(2023)Sheynin, Polyak, Singer, Kirstain, Zohar, Ashual, Parikh, and Taigman]{sheynin2023emu}
Shelly Sheynin, Adam Polyak, Uriel Singer, Yuval Kirstain, Amit Zohar, Oron Ashual, Devi Parikh, and Yaniv Taigman.
\newblock Emu edit: Precise image editing via recognition and generation tasks.
\newblock \emph{arXiv preprint arXiv:2311.10089}, 2023.

\bibitem[Tumanyan et~al.(2023)Tumanyan, Geyer, Bagon, and Dekel]{tumanyan2023plug}
Narek Tumanyan, Michal Geyer, Shai Bagon, and Tali Dekel.
\newblock Plug-and-play diffusion features for text-driven image-to-image translation.
\newblock In \emph{CVPR}, pages 1921--1930, 2023.

\bibitem[Wang et~al.(2023)Wang, Yu, and Zhang]{wang2023zero}
Yinhuai Wang, Jiwen Yu, and Jian Zhang.
\newblock Zero-shot image restoration using denoising diffusion null-space model.
\newblock In \emph{ICLR}, 2023.

\bibitem[Wu et~al.(2024)Wu, Chuang, Wang, Jia, Krishnakumar, Xiao, Liang, Yu, and Vajda]{wu2023fairy}
Bichen Wu, Ching-Yao Chuang, Xiaoyan Wang, Yichen Jia, Kapil Krishnakumar, Tong Xiao, Feng Liang, Licheng Yu, and Peter Vajda.
\newblock Fairy: Fast parallelized instruction-guided video-to-video synthesis.
\newblock \emph{CVPR}, 2024.

\bibitem[Wu et~al.(2023)Wu, Ge, Wang, Lei, Gu, Shi, Hsu, Shan, Qie, and Shou]{wu2023tune}
Jay~Zhangjie Wu, Yixiao Ge, Xintao Wang, Stan~Weixian Lei, Yuchao Gu, Yufei Shi, Wynne Hsu, Ying Shan, Xiaohu Qie, and Mike~Zheng Shou.
\newblock Tune-a-video: One-shot tuning of image diffusion models for text-to-video generation.
\newblock In \emph{ICCV}, 2023.

\bibitem[Xing et~al.(2023)Xing, Dai, Hu, Wu, and Jiang]{xing2023simda}
Zhen Xing, Qi Dai, Han Hu, Zuxuan Wu, and Yu-Gang Jiang.
\newblock Simda: Simple diffusion adapter for efficient video generation.
\newblock \emph{arXiv preprint arXiv:2308.09710}, 2023.

\bibitem[Yang et~al.(2023{\natexlab{a}})Yang, Mou, Yu, Wang, Meng, and Zhang]{yang2023neural}
Shuzhou Yang, Chong Mou, Jiwen Yu, Yuhan Wang, Xiandong Meng, and Jian Zhang.
\newblock Neural video fields editing.
\newblock \emph{arXiv preprint arXiv:2312.08882}, 2023{\natexlab{a}}.

\bibitem[Yang et~al.(2023{\natexlab{b}})Yang, Zhou, Liu, and Loy]{yang2023rerender}
Shuai Yang, Yifan Zhou, Ziwei Liu, and Chen~Change Loy.
\newblock Rerender a video: Zero-shot text-guided video-to-video translation.
\newblock In \emph{SIGGRAPH Asia 2023 Conference Papers}, pages 1--11, 2023{\natexlab{b}}.

\bibitem[Yang et~al.(2024)Yang, Zhou, Liu, and Loy]{yang2024fresco}
Shuai Yang, Yifan Zhou, Ziwei Liu, and Chen~Change Loy.
\newblock Fresco: Spatial-temporal correspondence for zero-shot video translation.
\newblock In \emph{Proceedings of the IEEE/CVF Conference on Computer Vision and Pattern Recognition}, pages 8703--8712, 2024.

\bibitem[Zhang et~al.(2023)Zhang, Mo, Chen, Sun, and Su]{Zhang2023MagicBrush}
Kai Zhang, Lingbo Mo, Wenhu Chen, Huan Sun, and Yu Su.
\newblock Magicbrush: A manually annotated dataset for instruction-guided image editing.
\newblock In \emph{Advances in Neural Information Processing Systems}, 2023.

\bibitem[Zhang et~al.(2024)Zhang, Wei, Jiang, ZHANG, Zuo, and Tian]{zhangcontrolvideo}
Yabo Zhang, Yuxiang Wei, Dongsheng Jiang, XIAOPENG ZHANG, Wangmeng Zuo, and Qi Tian.
\newblock Controlvideo: Training-free controllable text-to-video generation.
\newblock In \emph{The Twelfth International Conference on Learning Representations}, 2024.

\end{thebibliography}
}

\clearpage
\appendix

\section{Additional Implementation Details}
\label{appendix-section-implementation}

The evaluation was conducted using a collection of online resources and video clips from Panda-70M~\citep{chen2024panda70m}. \modelname{} can be applied to general image editing frameworks~\citep{hertz2022prompt, brooks2023instructpix2pix, fu2023guiding}. In this work, we used MGIE~\citep{fu2023guiding} as the base image editing model. We set the diffusion step $T$ to 10 and performed spatiotemporal adaptation through all cross-attention and self-attention layers. Our experiments showed that adaptation achieves the best performance when conducted on at least the first 8 steps.

We also observed that increasing the total diffusion step $T$ improves image detail but simultaneously raises the probability of artifacts. Through experimentation, we found that using a value between 5 and 10 generally yields good editing results while maintaining high processing speed. This balance ensures high-quality edits without introducing undesirable visual inconsistencies. For spatiotemporal adaptation, we collect attention variables from four frames.

\textbf{Test-time Editing Adaptation} is a process for refining the editing direction of the underlying model without relying on external data. The pipeline begins with an Edit \& Augment step, where a single frame is edited, and transformations are applied to both the source and edited frames to create a training set. Using this dataset, the underlying editing model is fine-tuned to adjust and improve the editing direction. We introduce the following transformations for each image pair, aimed at increasing variability while maintaining the structural integrity of the images: ($i$) slight rotation (up to ±5 degrees); ($ii$) translation (up to 5\% both horizontally and vertically); and ($iii$) after applying these transformations, cropping the images to between 75\% and 100\% of their original size to simulate changes in video sequence framing. Additionally, we apply shearing transformations of up to 10 degrees. These affine transformations introduce realistic variations, simulating the diversity of viewing angles typically encountered across different frames in a video. This approach helps the model adapt to the natural changes in perspective that occur during video sequences. For the tuning process, the training parameter for MGIE is the same as the tuning process of the underlying model. Specifically, we are using a learning rate of 5e-4 with AdamW optimizer, with a batch size of 16 and a total training of 200 steps. Our test-time adaptation process tunes the underlying image editing model towards a fixed editing direction. 
However, to the best of our knowledge, most video editing methods including the baselines used in this paper use an image generation or video generation model~\cite{wu2023tune, xing2023simda, guo2023animatediff}. One exception is one of our baselines, Fairy~\cite{wu2023fairy}, which uses an image editing model for video editing. However, since it did not open-source the code, it is hard to test the performance of test-time adaptation on other models. 

\textbf{Baseline Implementation} primarily follows the publicly available source code. For AnyV2V~\citep{ku2024anyv2v}, as it requires an edited first frame, we provide it with the first frame edited by \modelname{}. It inverts the source video into latent space and reconstructs the edited video using the edited frame as a condition. Rerender~\citep{yang2023rerender} edits the first frame using a diffusion model, modifies key frames, and interpolates the remaining frames based on the neighboring key frames. TokenFlow~\citep{tokenflow2023} inverts each video frame using DDIM to extract tokens and computes inter-frame correspondences via nearest-neighbor search. Keyframes are jointly edited at each denoising step to produce tokens, which are propagated across frames using pre-computed correspondences. The network replaces generated tokens with the propagated ones, iteratively refining the video into the final edited version. Video-P2P~\citep{liu2023video} employs a diffusion model with a shared unconditional embedding optimized for the reconstruction branch, while the initialized unconditional embedding is used for the editable branch, incorporating the editing instruction. Their combined attention maps generate the target video. Tune-A-Video~\citep{wu2023tune} uses a text-video pair as input and leverages pretrained T2I diffusion models for T2V generation. During fine-tuning, it updates the projection matrices in attention blocks with the standard diffusion training loss. At inference, it generates a new video by sampling latent noise inverted from the input video, guided by a modified prompt. For all methods requiring a new prompt rather than editing instructions, we use ChatGPT to rewrite the prompt. For Fairy~\citep{wu2023fairy}, as the code is not publicly available, we directly retrieved the video from their official website. For detailed configurations, please refer to their respective papers and open-source code.

From a high level, the difference between \modelname{} and other methods lies in three aspects:

\textbf{(i)} Other models do not consider the local editing process, meaning the editing may fail to faithfully follow the instruction across the entire frame. These methods typically rely on some attention-sharing mechanism without addressing the nuances of video editing.

\textbf{(ii)} For the information-sharing process across different frames, other approaches often directly share information without refinement, whereas \modelname{} employs \gatherswap{} to \textbf{emphasize consistency} in the shared information.

\textbf{(iii)} Their methods are often unsuitable for long videos due to limitations in the backbone architecture. In contrast, our global adaptation process \textbf{bypasses these limitations} in current models and hardware (e.g., GPU memory), enabling the editing of videos with up to a few thousand frames.

\section{Long Video Comparison}
\label{appendix:long-video-comparison}

\begin{figure}
    \centering
    \vspace{-2ex}
    \includegraphics[width=1.0\linewidth]{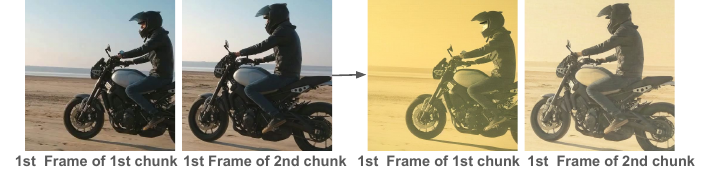}
    \caption{Editing results from two consecutive 5-second chunks. The editing instruction is ``Change the video to Japanese Woodprint painting.'' Even with the same model and random seed, the editing results can vary, leading to noticeable inconsistencies in the concatenated video.}
    \label{fig:chunk-edit}
    \vspace{-3ex}
\end{figure}

Since prior methods do not support long video editing, we divide long videos into 5-second segments, edit each segment separately, and then concatenate the results. \modelname{} significantly outperforms other baselines by a large margin. However, independently editing each chunk introduces noticeable inconsistencies. As an example shown in \cref{fig:chunk-edit}, applying AnyV2V~\cite{ku2024anyv2v} to two consecutive chunks results in visibly different editing effects across segments.

\begin{table*}[hbt]
\centering
    \caption{\textbf{Comparison with baselines using concatenated edited videos.} We evaluate our model against five previous open-source methods across three aspects. A `Tie' indicates comparable performance between models. Since prior methods do not support long video editing, we divide long videos into 5-second segments, edit each segment separately, and then concatenate the results.}
    \vspace{-2ex}
    \setlength{\tabcolsep}{2.0pt}
    \resizebox{\linewidth}{!}{
    \begin{tabular}{c|ccc|ccc|ccc|ccc|ccc}
        \toprule
        ~ & Ours & Rerender & Tie & Ours & TokenFlow & Tie & Ours & AnyV2V & Tie  & Ours & Video-P2P & Tie &
        Ours & Tune-A-Video & Tie\\
        \midrule
        Instruction Following ~ & \textbf{53.50} & 31.00  & 15.50 & \textbf{72.75} & 13.00 & 14.25  & \textbf{58.00} & 25.00 & 17.00 & \textbf{72.50} & 18.50 & 9.00 & \textbf{70.25} & 21.25 & 8.50\\
        Consistency ~ & \textbf{45.25} & 36.00 & 18.75 & \textbf{36.00} & 32.50 & 31.5 & \textbf{52.50} & 21.50 & 26.00 & \textbf{78.50} & 10.50 & 11.00 & \textbf{70.75} & 19.75 & 9.50\\
        Overall Quality ~ & \textbf{53.00}  & 27.00 & 20.00 & \textbf{70.75} & 15.50 & 13.75 & \textbf{72.50} & 13.25 & 14.25 & \textbf{61.75} & 14.75 & 23.50& \textbf{58.00} & 25.50 & 16.50 \\
        \bottomrule
    \end{tabular}
    }
    \label{table:human-evaluation}
\end{table*}

\section{Speed Analysis}

\modelname{} not only achieves great performance, but also offers impressive speed. The fine-tuning process takes approximately 1 minute, regardless of the video's length. For the global adaptation process, it takes InstructPix2Pix~\citep{brooks2023instructpix2pix} about 1 second per frame, and MGIE~\citep{fu2023guiding} around 3 seconds per frame. 

\textbf{Distribution Across GPUs:} Once we gather the frames, the editing for all frames can be performed on different GPUs simultaneously, as the frame editing process only depends on the fixed group frames. We utilize 8 GPUs for processing, which helps manage the load effectively.

\textbf{Total Processing Time for a 600-frame video:}
\begin{itemize}
    \item \textbf{MGIE:} 60 (fine-tuning) + $\frac{3 \times 600}{8} = 285$ seconds.
    \item \textbf{InstructPix2Pix:} 60 (fine-tuning) + $\frac{1 \times 600}{8} = 135$ seconds.
\end{itemize}

For the comparison with baselines, where only spatiotemporal adaptation is used (without fine-tuning or local adaptation), the time is:
\begin{itemize}
    \item \textbf{MGIE (without fine-tuning):} $\frac{3 \times 600}{8} = 225$ seconds.
    \item \textbf{InstructPix2Pix (without fine-tuning):} $\frac{1 \times 600}{8} = 75$ seconds.
\end{itemize}

\section{More Ablation Study}
\label{appendix:ablation-on-long-video}

In the main paper, we presented an ablation study on long videos. Here, we demonstrate the impact of various components of \modelname{} on videos less than 20 seconds in duration, where a dog rapidly moves its head and shakes its body. The provided editing instruction was "Change into a tiger." Our Local Latent Adaptation process effectively identifies the target area and performs precise edits. Our experiments also reveal that the initial edited frames largely determine the overall visual quality, as information from these root frames propagates throughout the entire video sequence. Test-time adaptation further ensures that the model adheres closely to the editing instructions.

In the absence of the \gatherswap{} process, relying solely on cross-frame attention results in inconsistencies across frames. Furthermore, while self-attention is commonly used to maintain frame consistency, we found that cross-attention significantly improves the quality of video editing. For example, when cross-attention is excluded, facial alignment with the source video is reduced, leading to less accurate transformations. In the right part of the experiment, we applied a style change to the video, transforming it into the aesthetic of a Japanese woodblock print. We observed that longer videos exhibit slightly lower visual performance compared to short ones, as minor mismatches can accumulate over a three-minute sequence with approximately 5,000 frames. We further conducted quantitative ablation on both long videos and short videos as in \cref{table:long_quantitative_ablation}.

\begin{figure*}[hbt!]
    \centering
    \includegraphics[width=\linewidth]{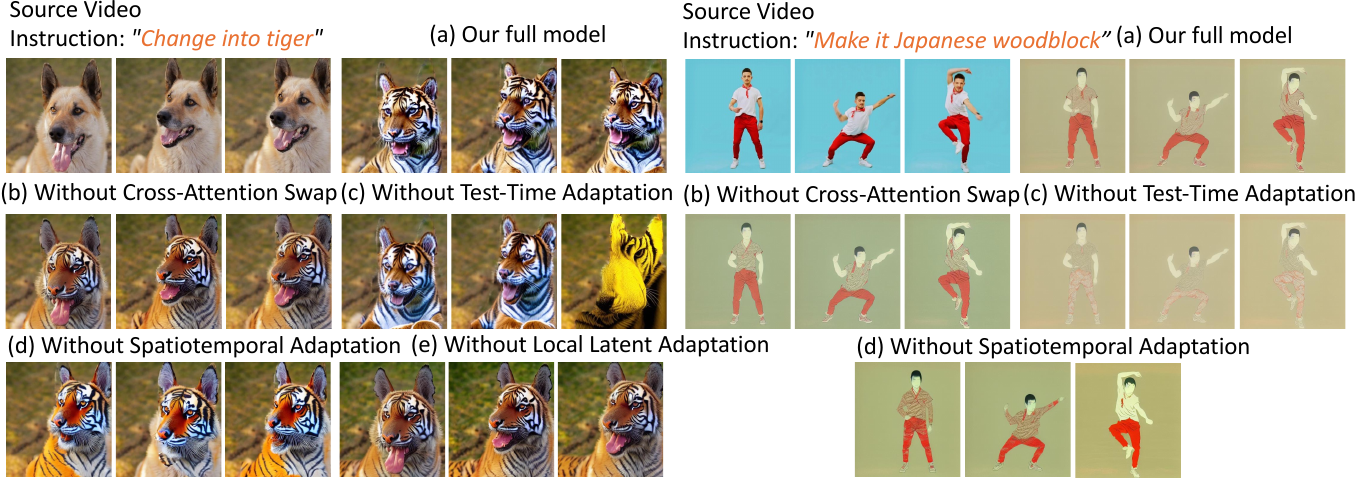}
    \caption{\textbf{Ablation study on videos less than 20 seconds.}}
    \label{fig:ablation-short}
\end{figure*}

\begin{table}[hbt]
    \vspace{-1.5ex}
    \caption{\textbf{Quantitative Ablation Study.} CA means Cross-Attention; TTA means Test-Time Adaptaion; SA means Spatiotemporal Adaptation; LLA means Local Latent Adaptation.}
    \vspace{-2ex}
\centering
\resizebox{1.0\linewidth}{!}{
    \begin{tabular}{cccccc}
        \toprule
        ~ & \modelname{} & w/o CA & w/o TTA &  w/o SA &  w/o LLA \\
        \midrule
        (Long) Frame-Acc $\uparrow$ ~ & \textbf{0.826} & 0.814 & 0.801  & 0.803 & 0.792 \\
        (Long) Tem-Con $\uparrow$ ~ & \textbf{0.942} & 0.923 & 0.913  & 0.909 & 0.910  \\
        \midrule
        (Short) Frame-Acc $\uparrow$ ~ & \textbf{0.869} & 0.852 & 0.844 & 0.842 & 0.833 \\
        (Short) Tem-Con $\uparrow$ ~ & \textbf{0.983} & 0.952 & 0.943  & 0.928 & 0.955 \\
        \bottomrule
    \end{tabular}
    \vspace{-3ex}
    }
    \label{table:long_quantitative_ablation}
\end{table}

\section{Analysis on Failure Cases}
\label{appendix:failure-cases}

\begin{figure*}
    \centering
    \includegraphics[width=\linewidth]{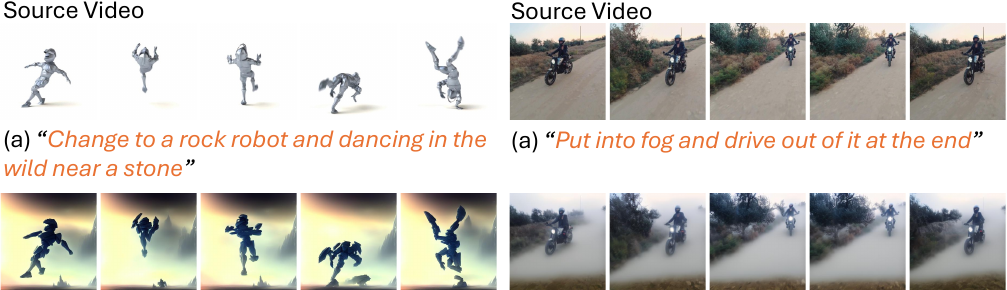}
    \caption{\textbf{Failure cases.} In the left example, a misalignment occurs during the interaction between the robot and the rock, despite accurately capturing the dance sequence. In the right example, while the driver is seamlessly integrated into the fog, the sequence fails to depict driving out process, leaving the edit incomplete.}
    \label{fig:failure-cases}
\end{figure*}

We highlight several failure cases where \modelname{} did not achieve the expected performance, as shown in \cref{fig:failure-cases}. The first challenge involves handling complex interactions. In the example on the left, while we successfully captured the intricate body dynamics during a sophisticated dance sequence, a misalignment occurred when the robot was supposed to interact with a rock, leading to inaccuracies at the point of contact. The second challenge relates to temporal dynamics. Although we seamlessly integrated the driver into the fog, the sequence did not show the car emerging from the fog, leaving the scene incomplete. In the future, we plan to incorporate more explicit temporal information into the editing process to better address these issues.

\section{Automatic Mask Generation}
\label{appendix:mask_generation}

\begin{figure*}
    \centering
    \includegraphics[width=\textwidth]{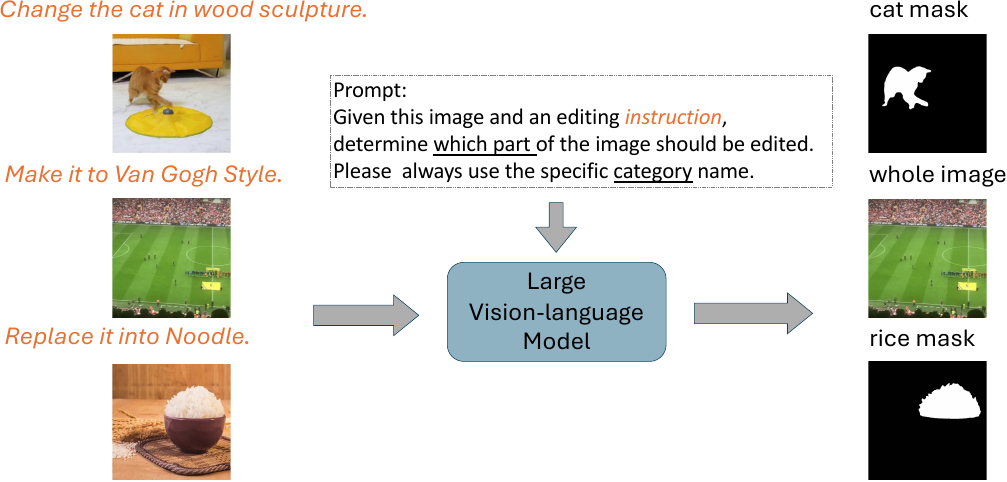}
    \caption{\textbf{Automatic mask generation.} A single frame from the video, along with a tailored text prompt encapsulating the editing instruction, is fed into a Large Vision-Language Model (LVLM), such as GPT-4, to generate a text description that specifies the region to be edited. If the designated editing area does not cover the entire image, this text description is then passed into a segmentation model, such as the Segment Anything model, to create a mask for the targeted region. This automated process allows for precise identification of the area to be modified, ensuring that only the relevant portion of the image is edited, while preserving the integrity of the rest of the frame.}
    \label{fig:mask-generation}
\end{figure*}

We present an automated mask generation pipeline aimed at enhancing user experience and streamlining the editing process, particularly for large-scale edits. Editing instructions often specify modifications to specific regions, but current end-to-end models tend to alter unintended areas. To address this, we designed an automated pipeline for mask generation, as illustrated in \cref{fig:mask-generation}.

First, a Large Vision-Language Model (GPT-4V in our experiment) is prompted to generate a textual description, \( P \), of the region to be modified for each frame. Using this description, we apply the Segment Anything model~\citep{Kirillov_2023_ICCV} to extract a mask that accurately delineates the target area for editing. It is important to note that we did not use GPT-4V during comparisons with baselines in the original paper.

In the optimal setting, VIA involves further tuning in the local adaptation process, which some baselines do not utilize. For fairness in comparisons, we degraded our model to use only Spatiotemporal Adaptation during all evaluations. This ensures that our results are directly comparable to baseline models without additional enhancements from local adaptation or the automated mask generation process.

\section{Performance on Other Backbone}
\label{appendix:performance_on_other_backbone}

\begin{figure}
    \centering
    \includegraphics[width=1.0\linewidth]{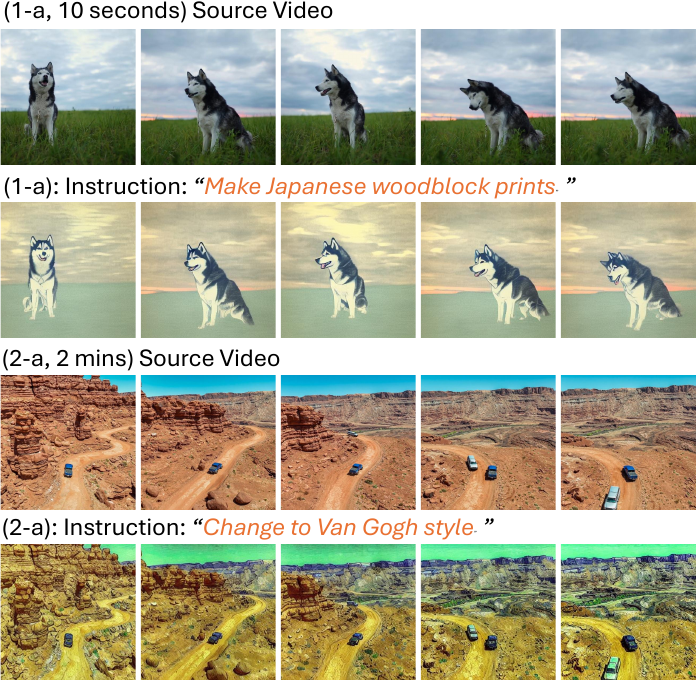}
    \caption{Editing results with InstructPix2Pix. The first one is a 10-second video, and the second one is a 2-minute video.}
    \label{fig:instruct-example}
\end{figure}

VIA can be equipped with various backbones. Here, we present the performance of another backbone, InstructPix2Pix~\cite{brooks2023instructpix2pix}. As shown in \cref{table:InstructPix2Pix-human-evaluation}, our model consistently outperforms baselines across multiple metrics. Compared to the MGIE backbone, VIA demonstrates improved \textit{Consistency} performance but slightly lower \textit{Instruction Following} performance. This aligns with the fact that MGIE incorporates an external instruction understanding module~\cite{liu2024visual}, which enhances its ability to handle complex editing instructions but diminishes the effect of shared group attention.
A similar trend is observed in \cref{table:InstructPix2Pix-automatic-evaluation}, where VIA achieves higher performance on \textit{Tem-Con} and \textit{Pixel-MSE} metrics but slightly lower performance on \textit{Frame-Acc}. Furthermore, VIA offers faster editing, as it bypasses the need for the additional instruction understanding process required by MGIE. Here for InstructPix2Pix, we used the same parameter setting as MGIE. In \cref{fig:instruct-example}, we present the results on both long and short videos.

\begin{table*}[hbt]
\centering
    \caption{\textbf{Human evaluation results.} We compare our model with five previous open-source methods from three aspects. `Tie' indicates the two models are on par with each other. Only spatiotemporal adaptation is used when compared with baseline models. Here we used InstructPix2Pix as the backbone. }
    \vspace{-2ex}
    \setlength{\tabcolsep}{2.0pt}
    \resizebox{\linewidth}{!}{
    \begin{tabular}{c|ccc|ccc|ccc|ccc|ccc}
        \toprule
        ~ & Ours & Rerender & Tie & Ours & TokenFlow & Tie & Ours & AnyV2V & Tie  & Ours & Video-P2P & Tie &
        Ours & Tune-A-Video & Tie\\
        \midrule
        Instruction Following ~ & \textbf{48.00} & 35.00  & 17.00 & \textbf{74.00} & 18.25 & 7.75  & \textbf{53.00} & 29.25 & 17.75 & \textbf{68.00} & 20.25 & 11.75 & \textbf{67.00} & 22.50 & 10.50\\
        Consistency ~ & \textbf{48.00} & 35.50 & 16.50 & \textbf{40.00} & 31.50 & 28.50 & \textbf{54.50} & 22.75 & 22.75 & \textbf{78.50} & 9.50 & 12.00 & \textbf{67.75} & 19.75 & 12.50\\
        Overall Quality ~ & \textbf{51.00}  & 28.00 & 21.00 & \textbf{59.75} & 23.25 & 17.00 & \textbf{61.75} & 31.50 & 6.75 & \textbf{60.25} & 24.25 & 15.50& \textbf{51.50} & 24.50 & 24.00 \\
        \bottomrule
    \end{tabular}
    }
    \label{table:InstructPix2Pix-human-evaluation}
\end{table*}

\begin{table*}[hbt]
    \caption{\textbf{Automatic evaluation results.} \modelname{} outperforms open-sourced video editing models in automatic metrics. Only spatiotemporal adaptation is used when compared with baseline models. Here we used InstructPix2Pix as the backbone. }
    \vspace{-2ex}
\centering
\resizebox{0.9\linewidth}{!}{
    \begin{tabular}{ccccccc}
        \toprule
        ~ & \modelname{} & Rerender & TokenFlow & AnyV2V &  Video-P2P & Tune-A-Video \\
        \midrule
        Frame-Acc $\uparrow$ ~ & \textbf{0.862} & 0.734 & 0.587 & 0.533 & 0.587 & 0.601\\
        Tem-Con $\uparrow$ ~ & \textbf{0.985} & 0.954 & 0.932  & 0.856 & 0.912 & 0.927 \\
        Pixel-MSE $\downarrow$ ~ & \textbf{0.010} & 0.016  & 0.018  & 0.026 & 0.020 & 0.019  \\
        Latency(sec) $\downarrow$  & \textbf{13} & 406 & 450 & 570 & 612 & 529 \\
        \bottomrule
    \end{tabular}
    \vspace{-2ex}
    }
    \label{table:InstructPix2Pix-automatic-evaluation}
\end{table*}

\section{Comparison on Attention Swapping Process}
\label{appendix:comparison-attention-process}

Attention variables within the U-net of diffusion models have proven to be highly correlated with the generated visual content and are widely used in various editing tasks~\citep{hertz2022prompt, cao2023masactrl, gu2023photoswap, liu2023video, ceylan2023pix2video}. In video editing, some methods train models to reconstruct the original videos and swap key attention features during the editing process~\citep{ku2024anyv2v, liu2023video}. Others suggest collecting attention variables independently from individual frame edits and applying them across frames~\citep{ceylan2023pix2video, wu2023fairy}; however, these independently generated attention variables often lack internal consistency. 

In contrast, our recursive \gath{} process ensures consistency within the attention group, which is especially crucial for long video generation, where maintaining coherence across thousands of frames is essential. Moreover, unlike previous methods that predominantly rely on self-attention, we also examine the significance of cross-attention layers, as highlighted in the ablation study.

\begin{figure}
    \centering
    \includegraphics{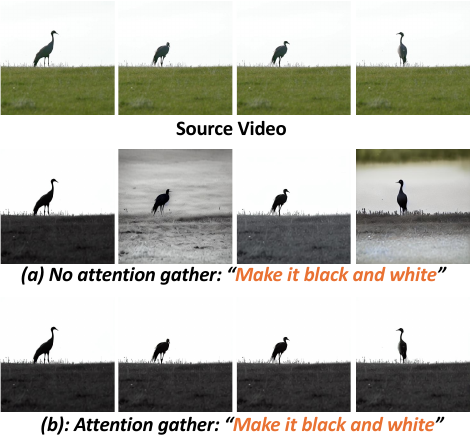}
    \caption{\textbf{The edited group frames with\&without attention gathering process.} The gathering process ensures in-group consistency, providing a fixed visual editing direction for all frames.}
    \label{fig:gather-analysis}
\end{figure}

Following the test-time adaptation process, each frame can be edited independently on separate GPUs during the spatiotemporal adaptation phase, significantly reducing the time required, particularly for long videos. We found that longer videos with more dynamics and scene changes benefit from a larger group size. In this work, we use a group size of 4 for all videos. The attention variable substitution process is performed throughout the entire denoising process, including the classifier-free guidance phase. The \gath{} process is essential to the model's success. As shown in \cref{fig:gather-analysis}, for the same video, using the same random seed and editing instruction, attention gathering produces much more consistent group frames. Without the gathering process, although each frame in the group still follows the instruction, they exhibit different semantic editing directions. With the gathering process, the group maintains internal consistency, and the attention variables from it provide stable guidance for all video frames in the subsequent editing process.

\section{Further Improvement with Better Root Frame}

\begin{figure}
    \centering
    \includegraphics[width=1.0\linewidth]{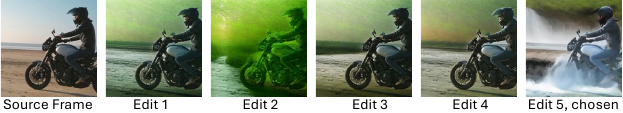}
    \caption{\textbf{Example of frame editing with different seeds.} Edited frames given the source frame on the left and editing instruction ``Driving on a river in a forest''}
    \label{fig:effect-of-choosing}
\end{figure}

In our practice, we observed that a high-quality root frame pair generally leads to improved performance, as illustrated in \cref{fig:effect-of-choosing}. In \cref{table:select-process}, we show that performance can be further enhanced by incorporating an additional selector. It is important to note that neither a human selector nor an automatic selector was used during the comparison with baselines. By selecting the optimal frame based on editing quality, we ensure that the best possible results are achieved without requiring complex video-level adjustments. This streamlined approach significantly enhances the effectiveness of our method and addresses concerns related to frame selection, allowing for more consistent and visually appealing edits across the video.

\begin{table*}[t]
\centering
    \caption{\textbf{The selection strategy further improves the results. }}
    \begin{tabular}{cccccc}
        \toprule
        ~ & Manuel & L1 & DINO & Random & No Test-time Adaptation \\
        \midrule
        Frame-Acc $\uparrow$ ~ & \textbf{0.891} & 0.882 & 0.887 & 0.873& 0.871\\
        Tem-Con $\uparrow$ ~ & \textbf{0.989} & 0.988 & 0.989 & 0.983 & 0.985\\
        Pixel-MSE $\downarrow$ ~ & \textbf{0.0102} & 0.0107 & 0.0108 & 0.0111 & 0.0113\\
        \bottomrule
    \end{tabular}
    \vspace{1ex}
    \label{table:select-process}
\end{table*}

\section{Blending Comparison}
\label{appendix:blending-comparison}

Our proposed Progressive Boundary Integration method differs significantly from traditional blending techniques by dynamically maintaining boundaries across both spatial and temporal dimensions in video editing. Unlike static methods that often cause artifacts like color bleeding or motion inconsistencies, it integrates inverted latent representations progressively, ensuring precise, localized edits without affecting non-targeted areas. The blending method commonly used in the diffusion process could be described as:

\begin{equation}
\label{equation:blending}
\bs{z}_{t}^{target} = \mask{} \cdot \bs{z}_{t}^{edit} + (1 - \mask{}) \cdot \bs{z}_{t}^{inverted}
\end{equation}

\begin{equation}
\label{equation:sample}
\bs{z}_{t-1}^{edit} = Sample(\bs{z}_{t}^{target}, \Phi, t)
\end{equation}

\begin{figure}
    \centering
    \includegraphics[width=1.0\linewidth]{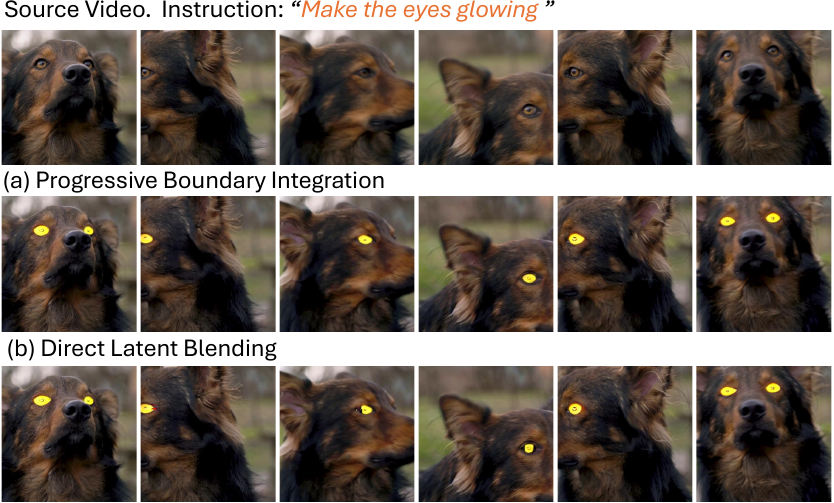}
    \caption{Comparison between Progressive Boundary Integration and direct latent blending reveals that the former achieves precise and consistent local editing results. For a closer examination, please zoom in on the eye area to observe the editing details.}
    \label{fig:blending-comparison}
\end{figure}

While this method works for individual frames, it fails to maintain consistent boundaries for dynamically changing objects in video sequences. This inconsistency leads to variations across frames in the editing area when replacing individual attention with group attention. In contrast, the dynamic mask defined in Equation 6 adjusts adaptively with each time step, allowing the attention to align more effectively with the target area as the diffusion process progresses.
In \cref{fig:blending-comparison}, we present examples of local editing applied to a dog's eyes with the instruction, “Make the eyes glowing.” Both Progressive Boundary Integration and direct latent blending successfully preserve the background. However, while the latter performs well on individual frames, it struggles with consistency across the video, as seen in the third frame from the left, where the glowing effect significantly shifts.
Experiments demonstrate that our method outperforms standard blending approaches, providing superior control and making it particularly well-suited for video edits that require preserving the integrity of unedited regions.

\section{Broader Impact} \label{appendix:broader-impact}

\modelname{} enhances video editing precision and efficiency, offering transformative benefits across multiple domains. In creative industries and education, it enables filmmakers, advertisers, and educators to produce high-quality, long-form content more efficiently. By reducing production costs and improving editing workflows, it allows for richer storytelling, clearer instructional videos, and more engaging educational materials.

Another key impact is the democratization of video editing. By simplifying advanced editing techniques, \modelname{} empowers non-professional users to create polished videos for social media, marketing, and personal projects. This expanded accessibility fosters greater creative expression while maintaining brand consistency and visual appeal in digital content.

While \modelname{} brings significant advancements, it also raises ethical and environmental considerations. The ability to seamlessly edit long videos introduces concerns about deepfakes and misinformation, highlighting the need for ethical safeguards and detection mechanisms. At the same time, its optimized processing reduces computational costs, promoting more sustainable video production.

Overall, \modelname{} has broad applications across industries, offering new creative possibilities while necessitating responsible and ethical implementation.

\section{Limitation}
\label{sec:limitation}

While \modelname{} has demonstrated impressive performance in video editing, it is not without limitations. Firstly, it inherits constraints from the underlying image editing model, which restricts the range of editing tasks to those predefined by the image model. For example, it is hard to achieve video motion-level editing if the backbone image editing model does not support it. Secondly, although \modelname{} performs well across a wide array of video editing tasks, its performance decreases when dealing with videos featuring complex interactions between objects. 
In the future, we plan to explore a more detailed part-to-part alignment to improve the model's capability in handling such scenarios.
\vspace{-1ex}

\end{document}